\documentclass{article}

\usepackage[preprint]{neurips_2026} 

\usepackage[utf8]{inputenc}    
\usepackage[T1]{fontenc}       
\usepackage{hyperref}          
\usepackage{url}               
\usepackage{booktabs}          
\usepackage{longtable}         
\usepackage{amsfonts}          
\usepackage{nicefrac}          
\usepackage{microtype}         
\usepackage{xcolor}            

\usepackage{graphicx}
\usepackage{multirow}
\usepackage{makecell}
\usepackage{tcolorbox}
\tcbuselibrary{breakable,skins,raster}
\usepackage{enumitem}
\usepackage{float}
\usepackage{threeparttable}
\usepackage{amsmath}
\usepackage{amssymb}
\usepackage{mathtools}
\usepackage{amsthm}
\usepackage[capitalize,noabbrev]{cleveref}

\definecolor{ingenix-purple}{HTML}{4F00B9}
\definecolor{ingenix-orange}{HTML}{FFAD00}

\tcbset{
  promptbox/.style={
    colback=gray!5,
    colframe=ingenix-purple!70,
    fonttitle=\bfseries,
    breakable,
    enhanced jigsaw,
    width=0.95\linewidth,
    boxsep=4pt,
    left=20pt,
    right=20pt,
    top=8pt,
    bottom=8pt,
    before skip=10pt,
    after skip=10pt,
  },
  reasoningbox/.style={
    colback=gray!5,
    colframe=ingenix-purple!70,
    fonttitle=\bfseries\footnotesize,
    fontupper=\scriptsize\raggedright,
    enhanced jigsaw,
    boxsep=2pt,
    left=4pt,
    right=4pt,
    top=3pt,
    bottom=3pt,
    boxrule=0.4pt,
    title style={fill=ingenix-purple!15},
    coltitle=black,
  }
}

\theoremstyle{plain}

\theoremstyle{definition}

\theoremstyle{remark}

\title{An explainable hypothesis-driven approach to Drug-Induced Liver Injury with \textsc{HADES}}

\author{%
  Maciej Wisniewski \\
  Ingenix.ai, Warsaw, Poland \\
  Laboratory of Bioinformatics and Computational Genomics, \\
  Warsaw University of Technology, Warsaw, Poland \\
  \And
  Bartosz Topolski \\
  Ingenix.ai, Warsaw, Poland \\
  \And
  Pawel Dabrowski-Tumanski \\
  Ingenix.ai, Warsaw, Poland \\
  Laboratory of Bioinformatics and Computational Genomics, \\
  Warsaw University of Technology, Warsaw, Poland \\
  \And
  Dariusz Plewczynski \\
  Ingenix.ai, Warsaw, Poland \\
  Laboratory of Bioinformatics and Computational Genomics, \\
  Warsaw University of Technology, Warsaw, Poland \\
  Laboratory of Functional and Structural Genomics, \\
  Centre of New Technologies, University of Warsaw, Warsaw, Poland \\
  \And
  Tomasz Jetka\thanks{Correspondence to Tomasz Jetka \texttt{<tomasz.jetka@ingenix.ai>}.} \\
  Ingenix.ai, Warsaw, Poland \\
  \texttt{tomasz.jetka@ingenix.ai} \\
}

\begin{document}

\maketitle

\begin{abstract}
Drug-induced liver injury (DILI) remains a leading cause of late-stage clinical trial attrition. However, existing computational predictors primarily rely on binary classification—a framing that limits generalization and yields no mechanistic insight to guide translational decisions. We argue that DILI prediction is fundamentally better posed as an explainable, hypothesis-generation problem. 

To support this paradigm shift, we introduce the DILER Benchmark, a novel dataset that extends beyond binary labels by augmenting a curated set of molecules with mechanistic hepatotoxicity hypotheses derived from biomedical literature. Furthermore, we present HADES, an agentic system designed to generate transparent, auditable reasoning traces. By fusing molecular-level predictions, metabolite decomposition, structural understanding, and toxicity pathways, HADES mechanistically assesses DILI risk. 

Evaluated on the DILER Benchmark, HADES outperforms existing models in binary classification, achieving a ROC-AUC of 0.68 on the Test Set and 0.59 on the challenging Post-2021 Set (compared to 0.63 and 0.50 for DILI-Predictor, respectively). More importantly, we establish a baseline for the novel task of mechanistic hypothesis generation, where HADES achieved a Hypothesis Alignment Fuzzy Jaccard Index of 0.16. This result underscores the inherent complexity of the task, while highlighting the critical need for advanced, explainable approaches in predictive toxicology.
\end{abstract}

\section{Introduction}\label{section:introduction}

Early identification of compounds likely to cause toxicity remains a major challenge in drug discovery and development across both preclinical and clinical stages. Among the organs most susceptible to drug-related adverse effects, the liver is of particular importance because it serves as the primary site of xenobiotic biotransformation, detoxification, and elimination, and is therefore one of the most frequent targets of adverse drug reactions, collectively referred to as drug-induced liver injury (DILI).

From a drug development perspective, DILI is responsible for approximately $13\%$ of clinical trial failures related to safety \citep{Ewart2022}, while hepatotoxicity represents the most common single cause of post-approval drug withdrawals, accounting for approximately $18-21\%$ of such cases \citep{Taylor2025}. Idiosyncratic DILI in particular typically escapes Phase I-II trials and surfaces only in larger populations during late-stage trials or after market approval. Translatability from preclinical models is also limited: roughly half of human hepatotoxic compounds fail to produce comparable signals in standard animal models, with liver-specific animal-to-human concordance substantially lower than for most other organs \citep{Leenaars2019}, and primary human hepatocytes reach only about $50\%$ sensitivity at 90-95\% specificity \citep{Chen2014}.

DILI is not a homogeneous phenomenon. It is divided into intrinsic DILI, which is typically dose-dependent and relatively predictable, and idiosyncratic DILI, which is rare and strongly shaped by host factors; phenotypically it presents as hepatocellular, cholestatic, or mixed. Its biological basis involves overlapping mechanisms -- reactive metabolite formation, oxidative stress, mitochondrial dysfunction, disruption of bile acid homeostasis, immune responses, and host genetic factors -- and clinical diagnosis remains one of exclusion in the absence of a single specific biomarker \citep{Hosack2023}. DILI is therefore better described as a family of mechanistically distinct causal scenarios than as a single prediction target, and actionable support for toxicological investigation requires reasoning about mechanism rather than only producing a binary risk label.

Existing computational approaches -- toxicophore rules, QSAR, classical machine learning, deep neural networks, graph-based models -- almost universally cast the problem as binary classification, with no clear winner and only moderate performance on real-world benchmarks. Beyond raw scores, such framing offers limited support for downstream decision-making: it does not explain why a compound is flagged, which mechanism is responsible, what evidence supports the hypothesis, or which uncertainties should be prioritized next. These shortcomings are compounded by benchmarks designed for classification rather than for evaluating mechanistic reasoning. Large language model (LLM) approaches, particularly agent-based architectures, are promising in this regime because they support decomposition into specialized steps such as molecular information extraction, metabolism analysis, identification of candidate mechanisms, literature retrieval, and synthesis of competing hypotheses with supporting evidence. Accordingly, in this work we i) introduce a DILI hypothesis benchmark and ii) an agentic framework designed not only to predict risk, but also to organize and mechanistically explain that risk in a form that is actionable for downstream investigation.

\section{Related Works}\label{section:related_works}

\subsection{Biological and Mechanistic Complexity of DILI}\label{section:related_works:biology}

Hepatotoxicity does not correspond to a single biological process. Mechanistic studies show substantial heterogeneity: structurally diverse compounds can induce similar clinical phenotypes through distinct pathways, while a single compound may engage multiple mechanisms simultaneously \citep{SkatRrdam2025}. This diversity has two implications for predictive modeling. First, a single binary hepatotoxicity label collapses multiple causal pathways, limiting the ability of models to distinguish mechanistically distinct compounds with similar outcomes. Second, practical utility requires models that can identify which pathways are likely engaged, provide supporting evidence, and account for alternative hypotheses -- motivating frameworks and methods that move beyond risk prediction toward explicit mechanistic reasoning.

\subsection{Adverse Outcome Pathways}\label{section:related_works:current:aop}

Among the most insightful frameworks for organizing mechanistic knowledge are Adverse Outcome Pathways (AOPs), curated through resources such as the AOP-Wiki \citep{AOPWiki}. An AOP represents how an initial biological perturbation propagates across levels of biological complexity until it culminates in an Adverse Outcome (AO): the pathway starts with a Molecular Initiating Event (MIE) and proceeds through one or more Key Events (KEs) to the AO, supporting relational graphs that encode stepwise causal reasoning. For DILI, hepatotoxicity rarely occurs as a single-step phenomenon - it reflects processes such as metabolic activation, oxidative stress, mitochondrial dysfunction, transporter inhibition, ER stress, or immune response - so AOPs supply a biologically grounded scaffold on which agentic reasoning can build, extend, and compare mechanistic hypotheses of DILI.




\subsection{Deep Learning Models for DILI Prediction}\label{section:related_works:current:models}

Computational prediction of hepatotoxicity has evolved from structural alerts and toxicophore rules toward data-driven modeling. Classical QSAR and machine-learning methods learned statistical associations between chemical features and DILI labels but were confined to chemical information. More recent deep models - message-passing graph neural networks and transformer-based molecular encoders - generally outperform descriptor baselines on standardized benchmarks, yet their outputs remain scalar risk scores devoid of mechanistic content. A complementary line of work injects biological information into the predictor itself: DILI-Predictor combines structural features with proxy labels from \textit{in vitro} and \textit{in vivo} assays \citep{Seal2024}, the multi-task framework of \citet{GarciadeLomana2025} exposes biologically meaningful intermediate signals, ToxReason~\citep{ToxReason} formulates toxicity prediction as a chain-of-thought task, Tox-Predictor~\citep{Bergen2025} incorporates gene expression data, and biomedical foundation models such as TxGemma~\citep{TxGemma} attach natural-language rationales to their predictions. Across all of these, mechanistic information is implicit - encoded in intermediate representations or auxiliary tasks - rather than expressed as explicit, structured hypotheses grounded in causal pathways.

\subsection{Benchmarks and Datasets for DILI Prediction}\label{section:related_works:benchmarks}

The quality of DILI models is tightly coupled to the benchmarks against which they are evaluated. Resources such as the DILI task within the Therapeutics Data Commons (TDC DILI) have been essential for standardizing classification evaluations and comparing structural and learned representations, and they remain well suited to the task they formalize: predicting a single binary or categorical hepatotoxicity label from structure. They do not, however, assess the quality of a mechanistic explanation, the evidence grounding of a hypothesis, the articulation of alternatives, or the ability to propose informative follow-up assays. A related concern is the underlying labels themselves: the literature has repeatedly noted contradictory annotations across studies, so that part of the apparent performance ceiling of existing models may reflect label noise as much as biological difficulty. Both observations point in the same direction—classification benchmarks are valuable but insufficient, and mechanistic reasoning deserves to be evaluated on its own terms.

\subsection{Agentic Systems}\label{section:related_works:agentic_systems}

We define an \emph{agentic system} as a large language model augmented with an explicit action interface, operating through iterative cycles of reasoning, tool use, evidence integration, and plan refinement. In biomedical contexts this paradigm has begun to mature: TxAgent~\citep{gao2025txagent} integrates a large library of FDA- and Open-Targets-derived tools for therapeutic reasoning, ToolUniverse~\citep{ToolUniverse} extends the idea into general infrastructure for composing tool workflows independently of the underlying model, and CellType~\citep{celltype_agent} demonstrates an industrial-scale instantiation spanning target discovery, compound profiling, and safety assessment. Our work adopts the same paradigm but specializes it to preclinical hepatotoxicity, with an explicit emphasis on integrating evidence across scales—from molecular features and AOP-based reasoning to organism-level outcomes—in the service of mechanistic hypothesis generation rather than open-ended biomedical question answering.

\section{Datasets}\label{section:datasets}

The datasets used in this work span multiple levels of abstraction: binary datasets capture only the presence or absence of hepatotoxicity, intermediate resources introduce mechanistic proxies such as molecular initiating events and pathway-level signals, and curated benchmarks encode full literature-grounded mechanistic hypotheses.


\subsection{DILI Dataset}\label{section:datasets:dili_dataset}

A wide range of classification strategies and preprocessing pipelines have been proposed for the DILI task. In this work, we adopt the previously processed dataset introduced in the DILI Predictor study~\citep{Seal2024}. This dataset was constructed by integrating the DILIst and DILIrank resources~\citep{Olubamiwa2025, Thakkar2020}, yielding a total of 1,111 unique compounds forming a so-called gold-standard DILI dataset.

We retain the same scaffold-based partition as DILI Predictor, comprising a training set ($n=888$) and an independent test set ($n=223$). This split is used for benchmark consistency and direct comparability with previously reported results, while molecular processing within \textsc{HADES} follows a separate pipeline, as discussed in Appendix~\ref{app:dili_preprocessing_note}.

In our evaluation, we utilize the 223 compounds from the held-out test set. In addition, we extend the analysis with a Post-2021 Benchmark consisting of 34 molecules not included in the original dataset. These compounds were manually curated based on the most recent available data and include small-molecule-like substances approved by the FDA since 01/01/2022 that are absent from existing DILI reference datasets.

\subsection{Mechanistic DILI Dataset}\label{section:datasets:mechanistic}

\subsubsection{Lomana dataset}

To support mechanistic characterization of drug-induced liver injury, we integrated the liver-related \textit{in vitro} endpoint collection reported by \citet{GarciadeLomana2025}. This resource comprises 28 endpoints related to liver toxicity and liver function, including transporter substrate and inhibition readouts, endoplasmic reticulum stress, and multiple forms of mitochondrial toxicity. We use these endpoints as \textit{proxy} mechanistic tasks within \textsc{HADES}, complementing compound-level DILI labels with biologically interpretable signals linked to molecular events and downstream cellular responses.

\subsubsection{Proxy MIE Datasets}\label{section:datasets:mechanistic:ours}

Although the Garcia de Lomana collection provides broad liver-relevant \textit{in vitro} coverage, it does not fully span the mechanistic scope required by \textsc{HADES}: several MIEs highlighted in the DILI literature \citep{Callewaert2024} are either absent or represented by assay formulations that do not match the perturbation type our framework needs (mismatch detailed in Appendix~\ref{app:mie_dataset}).


\begin{figure}[t]
\centering
\includegraphics[width=0.8\linewidth]{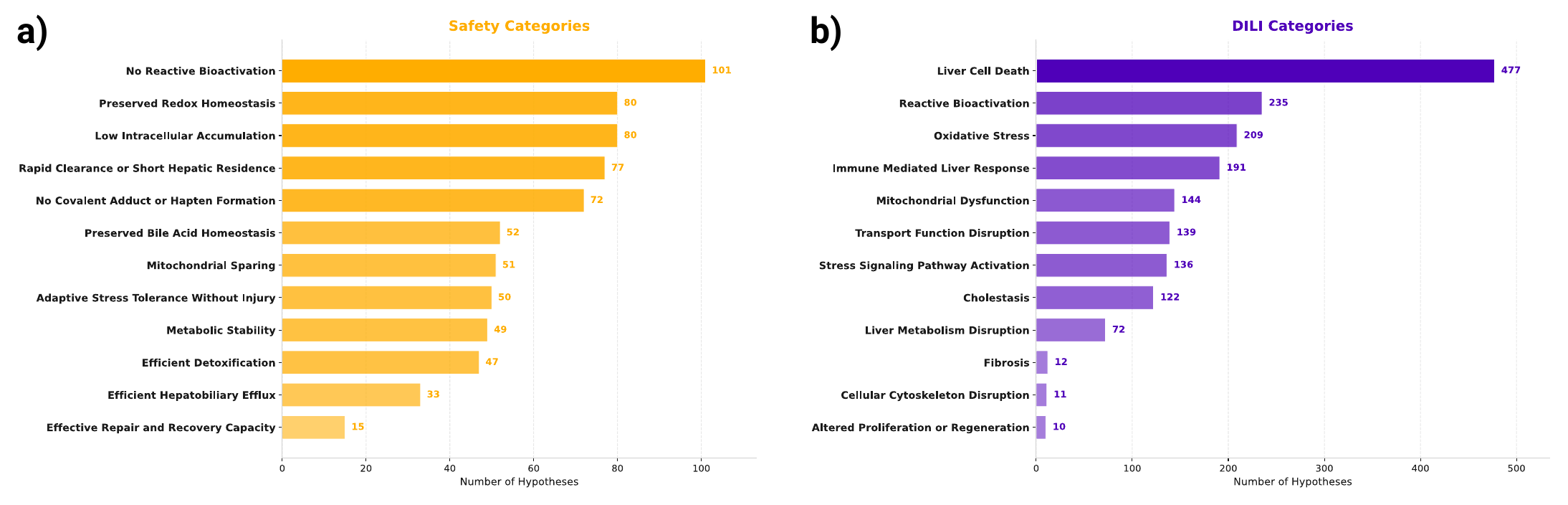}
\caption{The analysis of DILER Benchmark. Distribution of (a) Safety Categories and (b) Hepatotoxic Categories across the DILER Benchmark.}
\label{fig:diler_analysis}
\end{figure}

To address these gaps, we assembled an additional internal MIE-focused data collection, with curation details provided in Appendix~\ref{app:mie_processing}. Briefly, we integrated target-specific evidence from ChEMBL and EveBIO \citep{evebio}, standardized measurements into activation-, inhibition-, and binding-oriented endpoint classes, and harmonized them into binary target-level datasets using a common potency threshold. Labels were refined through compound--target consistency checks, conservative propagation of negative binding evidence, conflict reconciliation across sources, removal of compounds overlapping with the DILI benchmark, and filtering of highly imbalanced targets. The resulting collection is leakage-controlled and mechanistically focused, improving coverage of MIEs absent from, or insufficiently resolved by, the Garcia de Lomana resource.

A comparative overview of the AOPs considered in this study, their corresponding MIEs, the Garcia de Lomana coverage, and the additional endpoint coverage introduced by our internal collection is provided in Appendix~\ref{app:mie_dataset}.

\subsection{DILER Benchmark}\label{section:datasets:diler_benchmark}

Existing DILI benchmarks assess whether a model assigns the correct risk label, but rarely whether it can justify that label with a biologically meaningful mechanism -- a distinction that matters for toxicologists deciding which liabilities to investigate or which experiments to prioritize. The \textsc{DILER} Benchmark addresses this gap by centering on mechanistic hypotheses rather than endpoint labels. It covers the 223 molecules from the DILI Predictor test split and 34 additional compounds from our Post-2021 extension, with 929 curated hypotheses describing stepwise causal paths from compound-level properties to hepatotoxic or safety-relevant outcomes.

Each hypothesis is a compact mechanistic storyline of 5--7 reasoning steps, with a short title, an A--E hepatotoxicity label analogous to the LiverTox likelihood score, and one or more (non-exclusive) mechanistic categories; for most hypotheses we also provide a suggested biological assay. The benchmark was assembled with an agentic AI workflow using PubMed and clinical-trial evidence retrieval, followed by Deep Research and structured synthesis, with the goal of distilling available evidence into standardized, reviewable hypotheses that preserve causal logic. Category distributions are shown in Figure~\ref{fig:diler_analysis}; the full inventory with representative examples is in Appendix~\ref{app:diler_categories}, and hypothesis similarities in Appendix~\ref{app:diler_similarity}.

\section{Methods}\label{methods}

\subsection{HADES Structure}\label{section:methods:hades_structure}

\begin{figure}[t]
\centering
\includegraphics[width=0.7\linewidth]{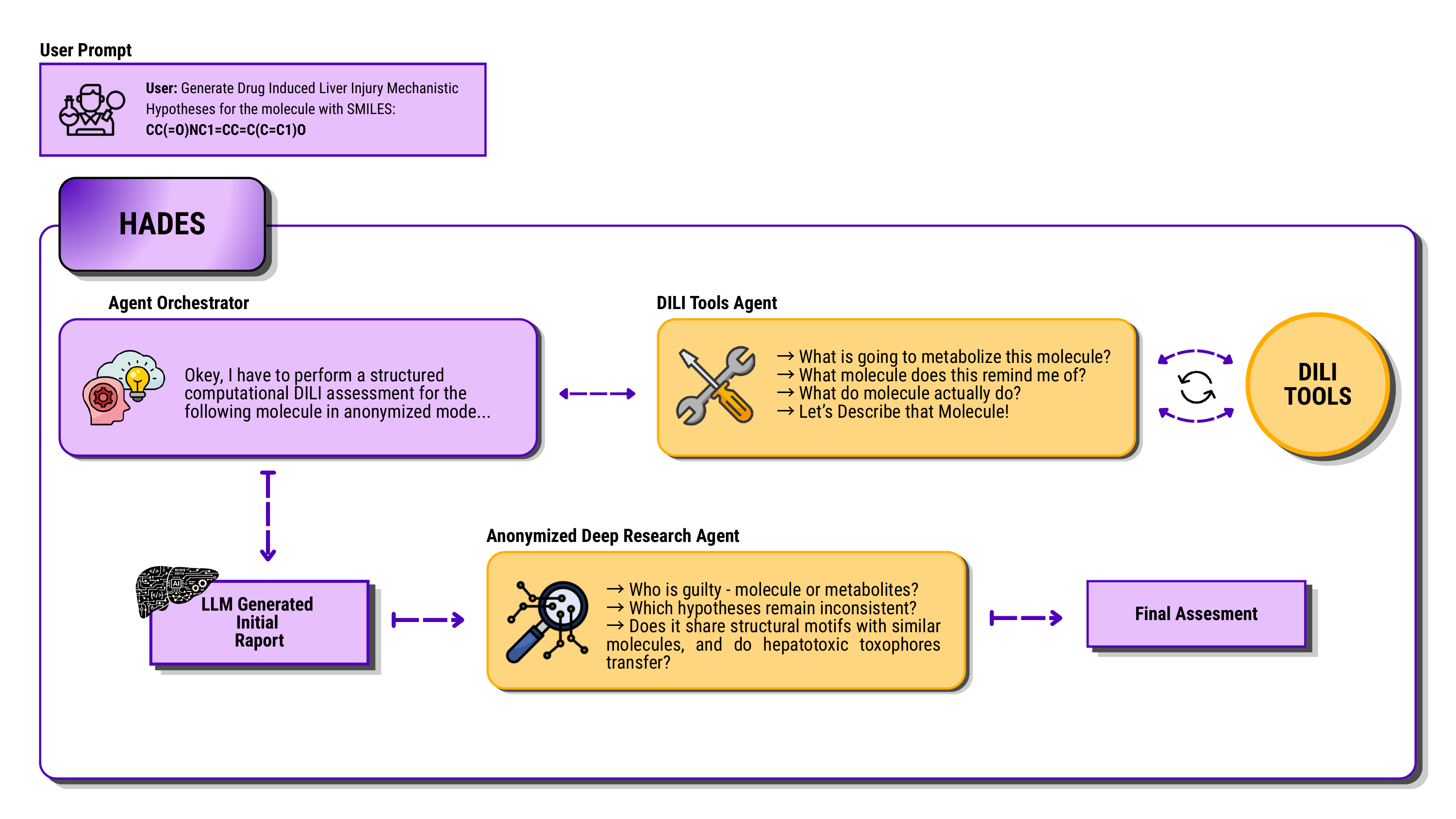}
\caption{The Structure of HADES agentic framework.}
\label{fig:hades_structure}
\end{figure}

\textsc{HADES} is an agentic pipeline that produces a structured DILI hypothesis landscape for a query molecule while preserving strict confidentiality of its chemical identity. It is implemented as a LangGraph State Machine~\citep{LangGraph} that delegates reasoning to two subagents executed in a fixed order: (i) the DILI subagent, which acts as an \emph{in-silico toxicologist} -- it iteratively decides which computational tools to invoke and synthesizes the outputs into a structured \emph{dossier} of molecular features, predicted liabilities, candidate metabolites, and structurally related compounds -- and (ii) DeepResearch, which is then invoked as a mandatory successor and confronts that dossier with the external biomedical literature, returning a cited argumentative report (Figure~\ref{fig:hades_structure}). We use Google's \textit{Gemini-3-Flash-Preview} as the deliberative backbone of the first stage.

The pipeline enforces an epistemic division of labour. The DILI subagent is the only component allowed to access the query molecule directly and interrogate it with predictive tools. DeepResearch, in contrast, may confront the emerging picture with external evidence but is forbidden from re-entering the computational toolbox. The first stage proposes; the second contextualizes; only after this contextualization is \textsc{HADES} permitted to formulate mechanism-level DILI hypotheses.

\textbf{Molecule anonymization at DeepResearch.} A central design principle in \textsc{HADES} is the strict anonymization of the studied molecule at the hand-off from the DILI subagent to DeepResearch. Only the DILI subagent ever sees the raw structural identifiers; downstream, the entity is referred to only as `the studied compound', and the molecule-level information propagated is limited to a tool-generated structural description and non-proprietary analogue identifiers from the internal similarity tool. This constraint blocks evidence contamination -- were DeepResearch given the exact identifier, it could retrieve literature describing the very molecule under evaluation and collapse a prospective assessment into post-hoc recall -- and enforces a clean separation of evidence classes: every claim in the final report is traceable either to direct computational output on the studied molecule, to indirect analogue evidence retrieved via DeepResearch, or to explicit mechanistic interpretation.


\subsection{Tools}\label{section:methods:tools}

\begin{figure}[t]
\centering
\includegraphics[width=0.7\linewidth]{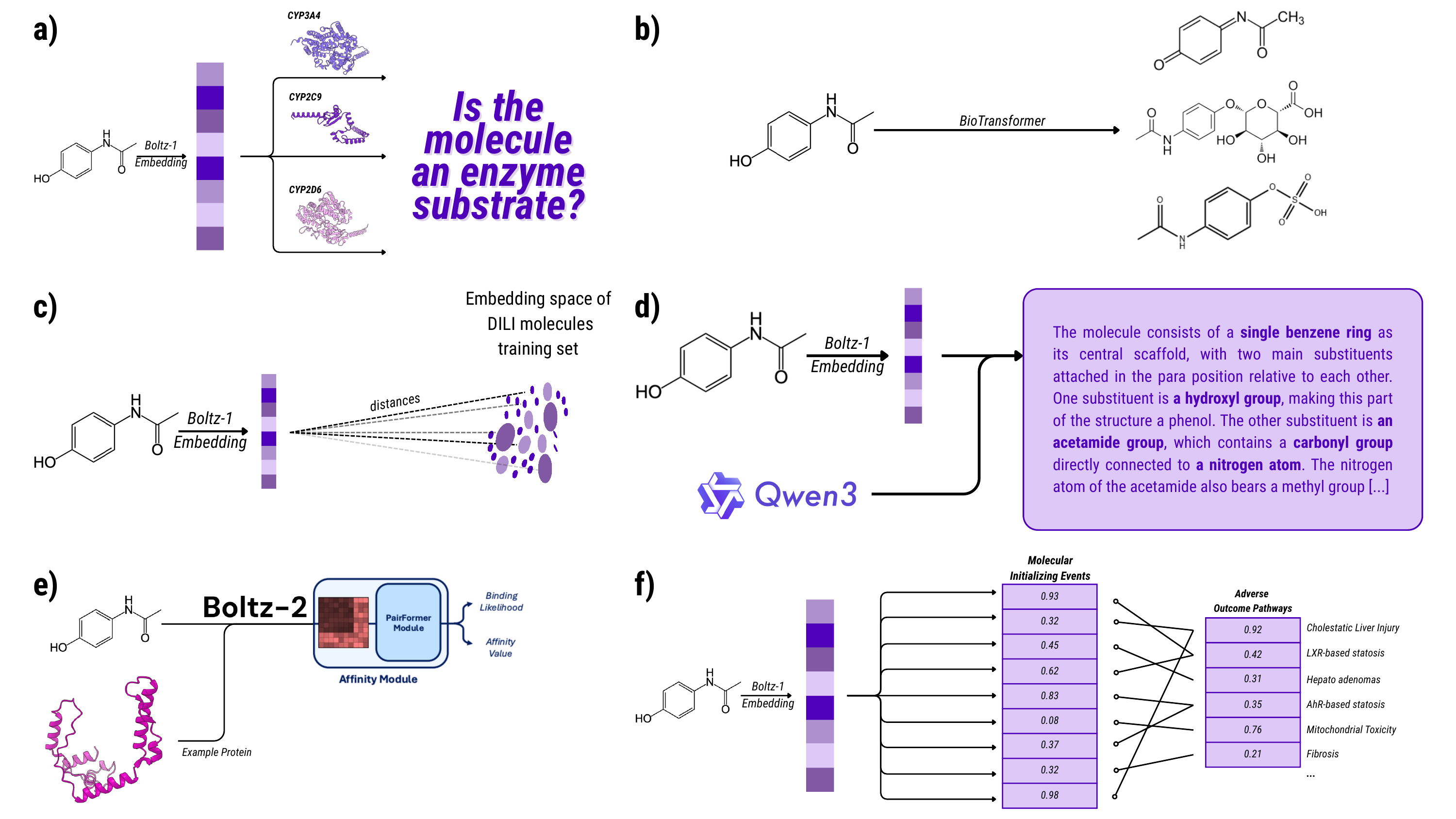}
\caption{Overview of the tool layer in \textsc{HADES}. (a) The Boltz-1 trunk embedding-based tool for predicting whether a given molecule is a substrate of cytochrome P450 (CYP) isoenzymes. (b) The Biotransformer tool for metabolites generation. (c) The Boltz-1 trunk embedding-based molecule similarity search for 888 Drug Induced Liver Injury labeled molecules. (d) The Molecule Structure Description Generator. (e) The Boltz-2 based tool for predicting binding affinity of protein - ligand complex. (f) The Boltz-1 trunk embedding-based tool for predicting MIEs and AOPs activation for a given molecule.}
\label{fig:tools_overview}
\end{figure}

Because DILI assessment requires reconciling mechanistic, structural, and metabolism-related signals, the DILI subagent is equipped with specialized tools that expose complementary views of the same compound (Fig.~\ref{fig:tools_overview}), so that hypothesis generation proceeds through explicit intermediate evidence rather than a single opaque toxicity score.

\subsubsection{AOP Predictor}\label{section:methods:tools:aop_predictor}

The AOP Predictor was designed to address a central limitation of conventional DILI modeling: a binary hepatotoxicity label is too coarse to support mechanistic reasoning. Because DILI emerges from a heterogeneous network of intracellular stress responses, transporter perturbations, and target-mediated effects, meaningful hypothesis generation requires access to intermediate biological signals rather than only a final risk score. We therefore introduced the AOP Predictor as a mechanistically oriented tool that estimates binary outcomes for the Molecular Initiating Events (MIEs) and Key Events (KEs) used in this study, together with additional protein targets that are not explicitly represented in AOP-Wiki but nonetheless contribute to hepatic homeostasis. In practice, these outputs serve as proxy cellular effects that help the agent construct, compare, and refine plausible mechanistic explanations for a given compound. Implementation details, including the Boltz-1-based backbone, uncertainty-aware reporting, and the Boltz-2 affinity-gating step, are given in Appendix~\ref{app:tool_implementation:aop_predictor}.

\subsubsection{Similarity Search}\label{section:methods:tools:similarity_search}

Similarity Search complements explicitly mechanistic tools in regions where current biology is incomplete or weakly formalized -- broader structural context, latent chemical liabilities, hapten-like effects, or host-modulated idiosyncratic responses that AOP- and target-centered predictors do not capture. The tool situates the query within a hepatotoxicity landscape defined by training-set neighbours and pairs the geometric retrieval with structured natural-language descriptions of both the query and its analogues, so that the second-stage agent can reason on shared scaffolds and recurring toxicophore motifs in a common vocabulary rather than on opaque distance scores. The aim is to enrich mechanistic reasoning with analogy, not to replace it; the entire view is constructed under full anonymization of the queried compound. Retrieval is performed over Boltz-1 trunk embeddings using a fractional-exponent energy distance between per-atom embedding sets, with each retrieved analogue and the query enriched by a \textsc{BoLEK}-generated structural description; the precise formulation is given in Appendix~\ref{app:tool_implementation:similarity_search}.

\subsubsection{Metabolite Tools}\label{section:methods:tools:metabolites}

In hepatotoxicity, the clinically relevant insult is often not the parent compound but a reactive or otherwise harmful metabolite produced through biotransformation, so treating the queried molecule as the only causal object -- as most DILI predictors do \citep{Seal2024, GarciadeLomana2025} -- is a default that quietly misses the actual driver. We therefore equip \textsc{HADES} with a metabolite-focused tooling layer that lets the agent reason over likely downstream metabolites and the MIEs they engage, in addition to the parent structure. This widens the evidential horizon from \emph{is this molecule hepatotoxic?} to \emph{which species along the metabolic trajectory is, and through what mechanism?}. The two-stage CYP-substrate gate plus BioTransformer-based metabolite enumeration is detailed in Appendix~\ref{app:tool_implementation:metabolites}.

\section{Results}\label{results}

\subsection{DILI Label Prediction}\label{section:results:binary_classification}

\begin{table}[h]
\centering
\small
\begin{threeparttable}
\caption{Binary DILI prediction on the DILI Test Set and DILI Post-2021 Benchmark.}
\label{tab:binary_classification}
\begin{tabular}{lcccccc}
\toprule
\textbf{Model} & \textbf{ROC-AUC} & \textbf{Bal Acc} & \textbf{MCC} & \textbf{Sensitivity} & \textbf{Specificity} & \textbf{F1} \\
\midrule
\multicolumn{7}{c}{\textbf{DILI Test Set (n=223)}} \\
\midrule
\textsc{HADES} & 0.68 & 0.58 & 0.18 & 0.81 & 0.36 & 0.78 \\
DILI Predictor & 0.63 & 0.59 & 0.16 & 0.61 & 0.56 & 0.65 \\
\midrule
\multicolumn{7}{c}{\textbf{DILI Test Set without TxGemma DILI Train Set (n=164)}} \\
\midrule
HADES & 0.65 & 0.54 & 0.09 & 0.77 & 0.31 & 0.74 \\
TxGemma-27B-Predict & 0.63 & 0.57 & 0.14 & 0.72 & 0.41 & 0.73 \\
TxGemma-27B-Chat\tnote{*} & 0.53 & 0.51 & 0.02 & 0.42 & 0.61 & 0.52 \\
\midrule
\multicolumn{7}{c}{\textbf{DILI Post-2021 Benchmark}} \\
\midrule
\textsc{HADES} & 0.59 & 0.53 & 0.15 & 1.0 & 0.05 & 0.63 \\
DILI Predictor & 0.50 & 0.53 & 0.07 & 0.80 & 0.26 & 0.59 \\
TxGemma-27B-Predict & 0.70 & 0.58 & 0.28 & 1.0 & 0.16 & 0.65 \\
TxGemma-27B-Chat\tnote{*} & 0.79 & 0.66 & 0.41 & 1.0 & 0.32 & 0.70 \\
\bottomrule
\end{tabular}
\begin{tablenotes}\footnotesize
\item[*] Metrics are biased by the Chat variant's tendency to anchor reasoning on a claimed drug name -- predominantly via confident misidentification; see Appendix~\ref{app:txgemma_leakage}.
\end{tablenotes}
\end{threeparttable}
\end{table}

By design, \textsc{HADES} does not emit a binary verdict. Instead, it returns an A--E severity label on a LiverTox-style scale~\citep{livertox}, which we convert into an ordinal continuous score for comparison with classical baselines: $A \mapsto 1.0$, $B \mapsto 0.75$, $C \mapsto 0.5$, $D \mapsto 0.25$, and $E \mapsto 0.0$. Because $59$ of the $223$ Test Set compounds also appear in the public corpus used to fine-tune both TxGemma variants, Table~\ref{tab:binary_classification} reports a leakage-controlled subset ($n=164$) alongside the full split; the Post-2021 Benchmark sits outside any TxGemma training cutoff.

\begin{figure}[h]
  \centering
  \includegraphics[width=0.5\linewidth]{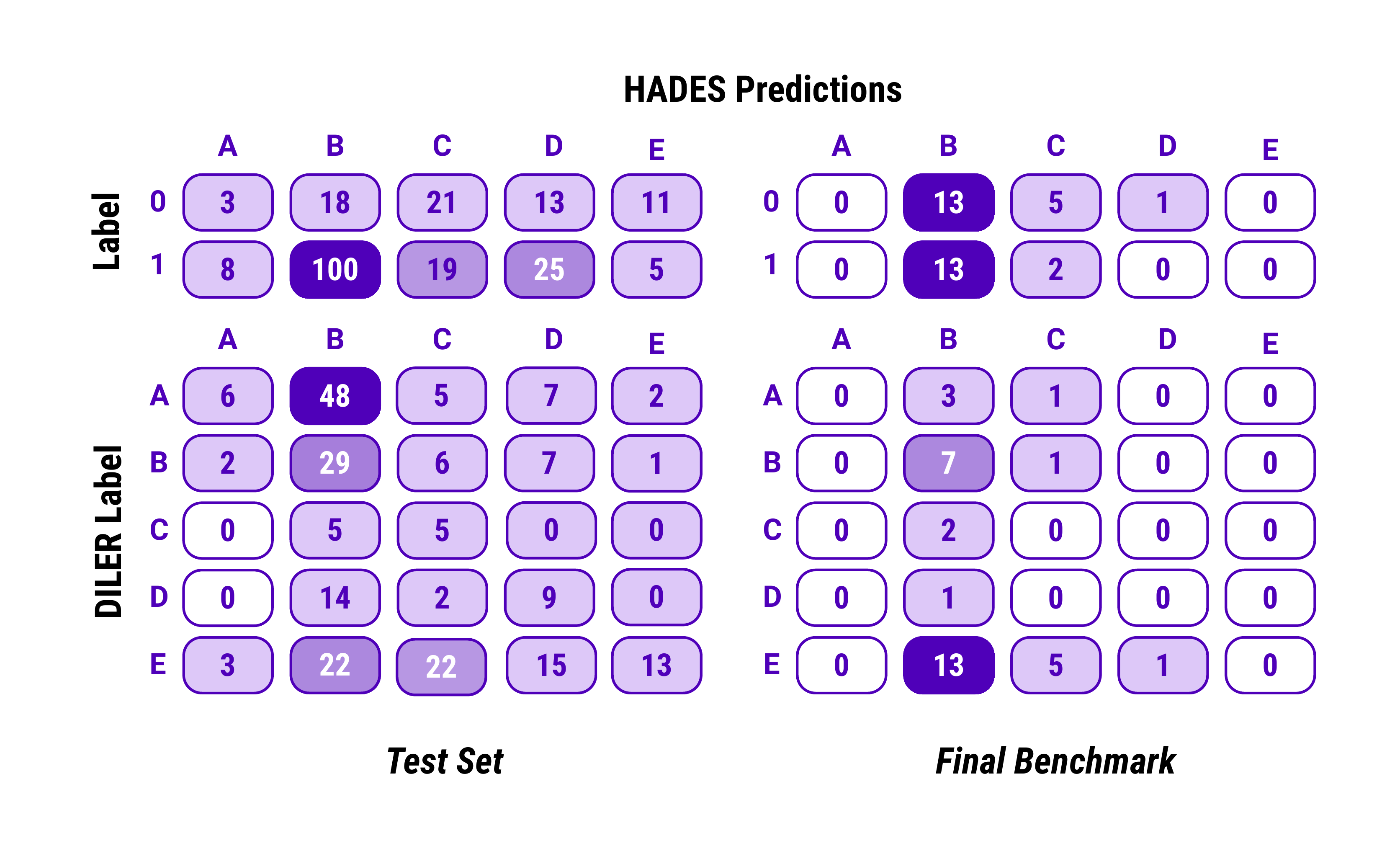}
  \caption{Dual confusion matrix on the Test Set and Post-2021 Benchmark for Binary and DILER Scale Labels.}
  \label{fig:dili_label_confusion}
\end{figure}

\textsc{HADES} retains non-trivial decision quality across all three views. It improves over DILI~Predictor on the full Test Set and on the Post-2021 Benchmark, and Figure~\ref{fig:dili_label_confusion} shows it distributes predictions more evenly across the A--E DILER scale than DILI~Predictor, which over-concentrates in the safest categories. On the leakage-controlled subset \textsc{HADES} leads \textsc{TxGemma-27B-Chat} on ROC-AUC metric and is on par with \textsc{TxGemma-27B-Predict}. The chat-tuned TxGemma's free-form reasoning routinely names the input molecule and then justifies its answer from the recalled drug profile rather than from structure. Verifying each claimed name against PubChem synonyms shows the dominant mode is not correct memorization but \emph{confident misidentification} (Appendix~\ref{app:txgemma_leakage}); on the Post-2021 Benchmark, where no claim survives this audit, the model's headline numbers reflect name-anchored heuristics rather than structural inference. \textsc{TxGemma-27B-Predict}, which exposes only a scalar score and so cannot be audited in the same way, posts the strongest aggregate ROC-AUC on Post-2021 ($0.70$ vs.\ \textsc{HADES}'s $0.59$). we read this honestly as a point in its favour, while noting that the Post-2021 split contains only $34$ compounds and is not statistically representative of out-of-distribution DILI behaviour, so single-split rankings on it should be interpreted with caution.

\subsection{Hypothesis Alignment}\label{section:results:hypothesis_alignment}

\begin{figure}[b]
    \centering
    \includegraphics[width=0.5\linewidth]{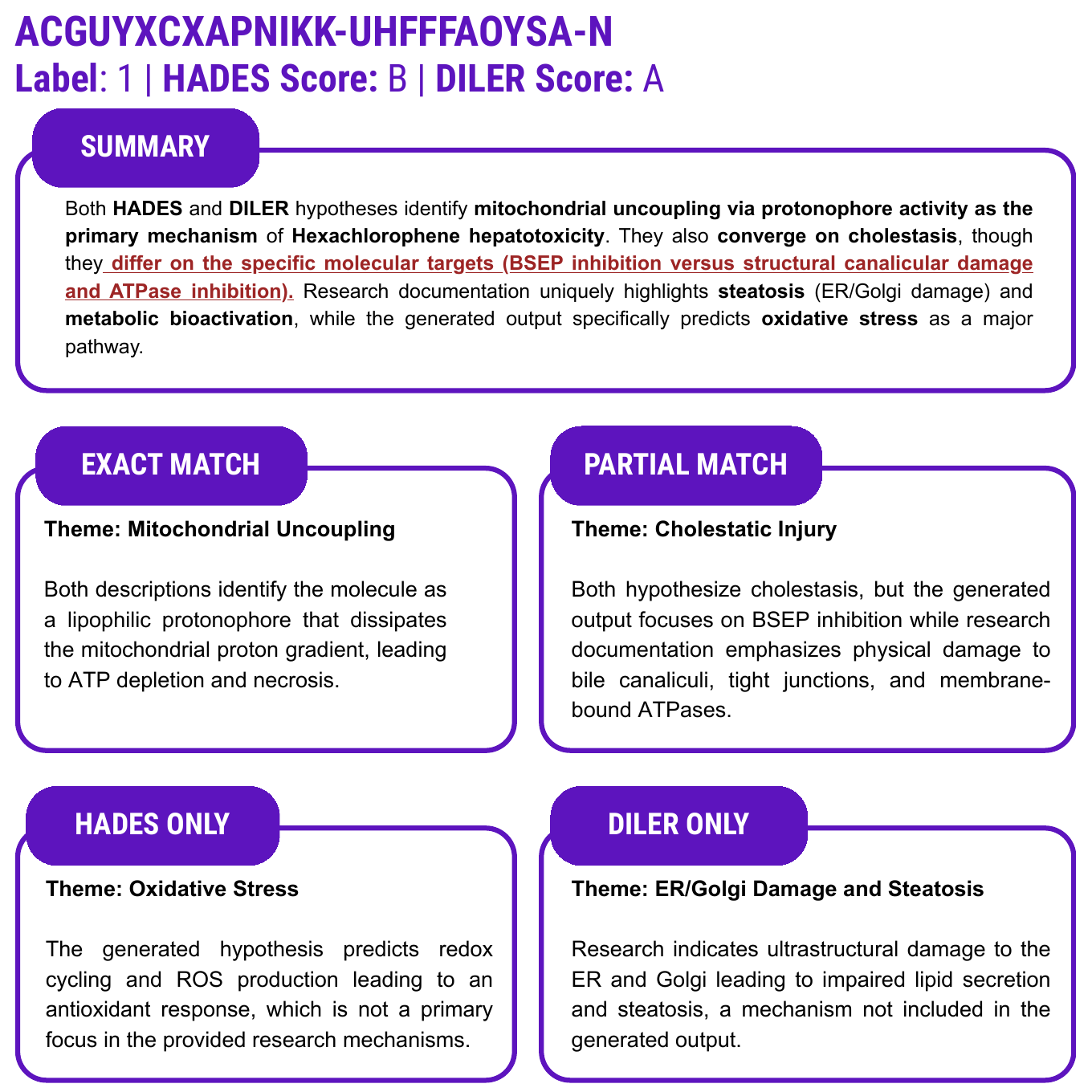}
    \caption{Example of the Structured Pairwise Alignment for a single compound.}
    \label{fig:hypothesis_alignment_example}
\end{figure}

Hypothesis Alignment asks whether the mechanistic story \textsc{HADES} told for a compound is the story a well-informed toxicologist would have told, by comparing the \textsc{HADES} hypothesis list against the DILER Benchmark reference. The procedure emits two complementary signals per compound: a continuous G-Eval~\citep{GEval} score in $[0,1]$ produced by Gemini-3-Flash-Preview as LLM-as-judge against a rubric that rewards matching mechanistic themes and causal direction while penalizing invented, missing, or contradicted mechanisms; and a \textbf{Structured Pairwise Alignment} that labels every \textsc{HADES}--DILER pair as \texttt{Exact Match}, \texttt{Partial Match}, \texttt{HADES Only}, \texttt{DILER Only}, or \texttt{Contradiction}. The per-pair labels are aggregated into set-similarity indices, mechanism-level Precision/Recall/F1, and error-mode rates (Contradiction, Hallucination, Miss). Definitions, the partial-match weighting, and full prompts are deferred to Appendices~\ref{app:hypothesis_alignment_metrics} and~\ref{app:evaluation_prompts}; a representative alignment is shown in Figure~\ref{fig:hypothesis_alignment_example}.


Table~\ref{tab:hypothesis_alignment} shows that alignment is partial rather than tight on both splits: strict set-similarity indices stay low while their fuzzy counterpart rises noticeably above them, reflecting many \texttt{Partial Match} pairs that share a mechanistic theme but disagree on scope or confidence. Hallucination is the dominant error mode, larger than Miss and Contradiction. Mirroring the binary setting, the head-to-head with \textsc{TxGemma-27B-Chat} runs on the leakage-controlled subset, where \textsc{HADES} leads on G-Eval while strict set-similarity, Precision, and Recall are essentially tied -- evidence that what \textsc{HADES} recovers is mechanism-relevant content rather than vocabulary overlap with TxGemma's training distribution. On Post-2021, \textsc{TxGemma-27B-Chat} reaches numerically higher set-similarity but lower G-Eval; since no recognition claim survives PubChem verification, this gap reflects the generic mechanistic vocabulary attached to confidently misidentified drugs rather than mechanism inference from structure. For the same reason -- internal recall confounding mechanistic inference -- we omit frontier closed-source LLMs from the headline tables; their evaluation, together with per-recognition alignment, is reported in Appendices~\ref{app:txgemma_leakage} and~\ref{app:frontier_llm_baselines}. Two illustrative \textsc{HADES}--DILER alignments are given in Appendix~\ref{app:hades_alignment_examples}.

\begin{table}[h]
\centering
\small
\makebox[\textwidth][c]{
\begin{threeparttable}
\caption{Hypothesis Alignment results across models on the DILER Test Set and Post-2021 Benchmark.}
\label{tab:hypothesis_alignment}
\begin{tabular}{lccccccccccc}
\toprule
\textbf{Model} 
& \textbf{G-Eval}
& \textbf{Jaccard} 
& \textbf{Dice}
& \textbf{Overlap}
& \makecell[l]{\textbf{Fuzzy} \\ \textbf{Jaccard}}
& \textbf{Precision} 
& \textbf{Recall}
& \textbf{F1}
& \makecell[l]{\textbf{Contr.} \\ \textbf{Rate}}
& \makecell[l]{\textbf{Halluc.} \\ \textbf{Rate}}
& \makecell[l]{\textbf{Miss} \\ \textbf{Rate}} \\
\midrule
\multicolumn{12}{c}{\textbf{DILER Test Set (n=223)}} \\
\midrule
HADES & 0.28 & 0.09 & 0.13 & 0.13 & 0.16 & 0.26 & 0.27 & 0.26 & 0.15 & 0.42 & 0.39 \\
\midrule
\multicolumn{12}{c}{\textbf{DILER Test Set without TxGemma DILI Train Set (n=164)}} \\
\midrule
HADES & 0.26 & 0.08 & 0.11 & 0.11 & 0.15 & 0.24 & 0.25 & 0.25 & 0.16 & 0.42 & 0.40 \\
TxGemma-27B-Chat\tnote{*} & 0.18 & 0.07 & 0.10 & 0.11 & 0.14 & 0.23 & 0.23 & 0.23 & 0.24 & 0.34 & 0.35 \\
\midrule
\multicolumn{12}{c}{\textbf{DILER Post-2021 Benchmark}} \\
\midrule
HADES & 0.20 & 0.08 & 0.12 & 0.13 & 0.12 & 0.19 & 0.21 & 0.20 & 0.24 & 0.45 & 0.36 \\
TxGemma-27B-Chat\tnote{*} & 0.17 & 0.10 & 0.14 & 0.15 & 0.15 & 0.25 & 0.26 & 0.25 & 0.22 & 0.34 & 0.37 \\
\bottomrule
\end{tabular}
\begin{tablenotes}\footnotesize
\item[*] Metrics are biased by the Chat variant's tendency to anchor reasoning on a claimed drug name - predominantly via confident misidentification on both splits; see Appendix~\ref{app:txgemma_leakage}.
\end{tablenotes}
\end{threeparttable}%
}
\end{table}

\section{Discussion}\label{discussion}

In DILI, a binary label alone is insufficient for decision-making, motivating the design of \textsc{HADES} as a system that jointly produces calibrated risk predictions and structured mechanistic rationales. On the held-out Test Set and the Post-2021 Benchmark, \textsc{HADES} improves over DILI~Predictor on ROC-AUC while, crucially, returning compact sets of explicit, traceable hypotheses with stepwise causal structure. It recovers dominant class-level mechanisms -- reactive bioactivation, mitochondrial dysfunction, transporter perturbation, and hapten-driven immunity -- with correct directionality.  This capability is enabled and evaluated through \textsc{DILER}, benchmark of literature-grounded mechanistic hypotheses, which allows assessment of whether predictions are correct \emph{for the right reason} rather than only numerically accurate. Additionally, we understand the limitations of the Deep Research synthesis and we are working on human curation of these hypotheses.

Our head-to-head against TxGemma highlights why mechanism-level evaluation matters. The chat-tuned variant routinely identifies the input molecule by name and reasons from the recalled drug profile rather than from structure; on splits where this name claim does not survive a PubChem audit, its apparent advantage on classification and set-similarity reflects name-anchored heuristics rather than prospective inference. \textsc{HADES}, in contrast, is anonymized at the hand-off to its literature stage, so its hypotheses are derivable from tool evidence and analogue context -- and it leads TxGemma on the LLM-as-judge G-Eval score, which weighs mechanistic substance over vocabulary overlap.

At the same time, evaluation exposes clear limitations of \textsc{HADES} itself. Hallucination remains the dominant error mode, particularly for compound-specific mechanisms, reflecting constrained AOP and MIE coverage, limited metabolite modelling, and under-representation of immune- and host-mediated pathways -- further compounded by anonymization, which by design restricts access to compound-specific evidence -- so \textsc{HADES} should remain decision support for human toxicologists rather than a standalone safety verdict. We are also mindful of the dual-use risk highlighted by \citet{Urbina2022}: toxicity-prediction systems can in principle be repurposed to optimize for harm, so \textsc{HADES} is positioned only to filter hepatotoxic liabilities out of drug pipelines, and we do not release weights or fine-tuned tool checkpoints that would lower the barrier to misuse. These limitations point to concrete extensions: richer mechanism predictors, broader metabolite tooling, and stronger consistency constraints between the computational dossier and the literature stage.

\section*{Software and Data}
The DILER Benchmark will be made publicly available upon release of the paper.

\bibliographystyle{plainnat}
\bibliography{references}

\newpage
\appendix

\section{Molecule Preprocessing}\label{app:dili_preprocessing_note}

For evaluation on the DILI Predictor benchmark, we intentionally preserved the original preprocessing-to-split pipeline in order to maintain strict comparability with the published study. Changing the preprocessing before splitting could alter scaffold assignments and, consequently, modify the train-test partition itself. For that reason, the benchmark experiments reported in this work use the same molecule preprocessing assumptions that underlie the DILI Predictor split.

At the same time, this preprocessing is not used as the default representation pipeline in \textsc{HADES}. In our assessment, some of the processed structures produced by that workflow are converted into uncommon resonance or charge-separated SMILES forms. Although such representations may be formally valid, they are often less standard from the perspective of downstream cheminformatics tools and less interpretable for mechanistic analysis. In particular, we observed that this preprocessing can lead to rare resonance-like SMILES variants that are not ideal for tool integration and agent-based reasoning.

For this reason, \textsc{HADES} uses a separate molecular preprocessing procedure for its internal workflow, aimed at generating chemically more standard and operationally more stable representations. We therefore distinguish between two objectives: preserving exact compatibility with the DILI Predictor benchmark split for evaluation, and using a different preprocessing strategy for molecules processed within \textsc{HADES} itself.

\section{Molecular Initiating Events Dataset Processing}\label{app:mie_processing}

To construct target-specific datasets for MIE modeling, we integrated activity information from two complementary sources, ChEMBL and EveBIO, and converted these records into harmonized binary classification tables. The overall workflow consisted of five stages: extraction of target-centered measurements from ChEMBL, assignment of pharmacological meaning based on endpoint type, target-level consistency curation and label augmentation, reconciliation with EveBIO, and final filtering designed to prevent data leakage and exclude datasets with insufficient class balance.

The workflow began with retrieval of bioactivity records from a local ChEMBL instance. For each protein target of interest, we queried standardized quantitative measurements linked to compounds with a defined molecular structure and retained only records associated with recognized potency-oriented endpoints. In practice, the extraction was restricted to $IC_{50}$, $EC_{50}$, and $K_d$ measurements, as these assay types provide a relatively consistent basis for downstream interpretation. In addition to the numerical activity value, we preserved the surrounding assay context, including assay identifiers, assay descriptions, confidence scores, compound identifiers, activity comments, and canonical SMILES representations.

Each measurement type was then mapped to a specific pharmacological interpretation. $IC_{50}$ values were treated as evidence of inhibition, $EC_{50}$ values as evidence of activation, and $K_d$ values as evidence of binding affinity. For all three endpoint types, we applied a uniform potency threshold of $10{,}000$ nM under the conventional assumption that lower values indicate stronger target engagement. Measurements below this threshold were assigned a positive label, whereas measurements greater than or equal to this threshold were assigned a negative label. This procedure yielded endpoint-specific binary datasets for inhibition, activation, and binding affinity at the level of individual protein targets.

After binarization, the resulting tables were deduplicated and subjected to target-level consistency control. In particular, when the same compound-target pair was labeled as positive in both the activation and inhibition datasets, the corresponding records were removed from both datasets because such assignments are not mechanistically consistent within the scope of this binary formulation. We subsequently performed logical label augmentation between the activation and inhibition datasets: a positive activation label implied a negative inhibition label for the same compound-target pair, and conversely, a positive inhibition label implied a negative activation label. This step increased label completeness while preserving directional consistency between mutually exclusive functional outcomes.

Binding affinity data were used as an additional source of negative evidence. Specifically, when a compound-target pair received a negative label in the $K_d$-based binding dataset, the same pair was also assigned negative labels in the corresponding activation and inhibition datasets. We did not propagate positive binding labels to functional datasets, because detectable target binding alone does not determine whether the interaction results in activation or inhibition. In this way, binding data were used conservatively to strengthen negative evidence without introducing unsupported assumptions about functional directionality.

The curated ChEMBL-derived datasets were then reconciled with EveBIO after protein target mapping between the two resources. For targets represented in EveBIO, we incorporated the corresponding records because EveBIO provides additional assay coverage for compound activity and antagonism. When discrepancies were identified between ChEMBL-derived labels and EveBIO labels for the same compound-target relationship, the conflicting ChEMBL record was removed and the EveBIO annotation was retained. This choice ensured that each final dataset contained only a single resolved label for a given compound-target pair.

To prevent information leakage into downstream benchmark experiments, we removed from every target-specific MIE dataset all compounds appearing anywhere in the DILI collection, regardless of whether they belonged to the training, test, or Post-2021 Benchmark. Finally, we retained only those target-specific datasets that satisfied minimum quality criteria for binary modeling, namely a majority class comprising less than $90\%$ of the dataset and a minority class containing more than 20 samples. The resulting collection therefore consisted of curated, leakage-controlled, and sufficiently balanced target-level MIE datasets suitable for subsequent mechanistic prediction tasks.

\section{Proxy MIE Datasets Overview}\label{app:mie_dataset}

\begin{table}[h]
\centering
\caption{Comparison of Adverse Outcome Pathway, where AOP and MIE reference their IDs from AOP-Wiki, and ChEMBL ID refers to the target identifier from ChEMBL.}
\label{app:tab:aop_mie_comparison}
\small

\begin{minipage}[t]{0.49\textwidth}
\centering
\setlength{\tabcolsep}{3pt}
\renewcommand{\arraystretch}{0.9}
\resizebox{\linewidth}{!}{%
\begin{tabular}{lllll}
\toprule
\textbf{AOP} & \textbf{MIE} & \textbf{Protein Target} & \textbf{ChEMBL ID} & \textbf{Mechanism} \\
\midrule

AOP 27 & MIE 41 & BSEP & 6020 & Inhibition \\

AOP 32 & MIE 147 & NOS2 & 4481 & Inhibition \\

\multirow{2}{*}{AOP 34}
& MIE 167 & LXR $\alpha$ & 2808 & Activation \\
& - & RXR $\alpha$ & 2061 & Activation  \\

\multirow{3}{*}{AOP 36}
& MIE 231 & PPAR $\alpha$ & 239 & Inhibition \\
& MIE 232 & PPAR $\beta$/$\delta$ & 3979 & Inhibition \\
& MIE 233 & PPAR $\gamma$ & 235 & Inhibition \\

AOP 37 & MIE 227 & PPAR $\alpha$ & 239 & Activation \\

AOP 41 & MIE 165 & AHR & 3201 & Activation \\

AOP 57 & MIE 18 & AHR & 3201 & Activation \\

\multirow{4}{*}{AOP 58}
& MIE 456 & CAR & 5503 & Inhibition \\
& MIE 468 & PPAR $\alpha$ & 239 & Inhibition \\
& MIE 167 & LXR $\alpha$ & 2808 & Activation \\
& - & RXR $\alpha$ & 2061 & Activation \\

AOP 59 & MIE 461 & HNF4A & 5398 & Inhibition \\

AOP 60 & MIE 245 & PXR & 3401 & Activation \\

\multirow{2}{*}{AOP 61}
& MIE 478 & NFE2L2 & 1075094 & Activation  \\
& MIE 479 & NR1H4 & 2047 & Activation \\

\multirow{2}{*}{AOP 62}
& KE 484 & AKT2 & 2431 & Activation \\
& KE 457 & SREBF1 & 4630818 & Activation \\

AOP 107 & MIE 715 & CAR & 5503 & Activation \\

AOP 108 & MIE 724 & PDK2 & 3861 & Inhibition \\

AOP 117 & MIE 25 & AR & 1871 & Activation \\

AOP 130 & MIE 828 & PLA2G1B & 4426 & Inhibition \\

AOP 220 & MIE 1391 & CYP2E1 & 5281 & Activation \\

AOP 318 & MIE 122 & NR3C1 & 2034 & Activation \\

AOP 383 & MIE 1740 & ACE2 & 3736 & Inhibition \\

AOP 494 & MIE 18 & AHR & 3201 & Activation \\

AOP 517 & MIE 239 & PXR & 3401 & Activation \\

\multirow{2}{*}{AOP 518}
& MIE 167 & LXR $\alpha$ & 2808 & Activation \\
& - & RXR $\alpha$ & 2061 & Activation \\
\bottomrule
\end{tabular}%
}
\end{minipage}
\hfill
\begin{minipage}[t]{0.49\textwidth}
\centering
\setlength{\tabcolsep}{3pt}
\renewcommand{\arraystretch}{0.9}
\resizebox{\linewidth}{!}{%
\begin{tabular}{lllll}
\toprule
\multicolumn{5}{c}{\textbf{Non Protein -- AOP Mechanisms}} \\
\midrule
\textbf{AOP} & \textbf{MIE / KE} & \multicolumn{2}{c}{\textbf{Effect}} & \textbf{Mechanism} \\
\midrule

AOP 144 & KE 177 & \multicolumn{2}{l}{\makecell[l]{Mitochondrial \\ dysfunction}} & Increase \\

AOP 273 & KE 1545 & \multicolumn{2}{l}{\makecell[l]{Mitochondrial \\ oxidative \\ phosphorylation}} & Decrease \\

AOP 328 & KE 1770 & \multicolumn{2}{l}{\makecell[l]{Mitochondrial \\ membrane \\ potential}} & Decrease \\

AOP 362 & KE 177 & \multicolumn{2}{l}{\makecell[l]{Mitochondrial \\ dysfunction}} & Increase \\

AOP 387 & KE 1770 & \multicolumn{2}{l}{\makecell[l]{Mitochondrial \\ membrane \\ potential}} & Decrease \\

AOP 454 & MIE 2017 & \multicolumn{2}{l}{\makecell[l]{Endoplasmic \\ reticulum \\ stress}} & Increase \\

\midrule
\multicolumn{5}{c}{\textbf{Protein -- Non AOP Mechanisms}} \\
\midrule
\multicolumn{2}{c}{\textbf{Target Protein}} & \multicolumn{2}{c}{\textbf{ChEMBL ID}} & \textbf{Mechanism} \\
\midrule
\multicolumn{2}{l}{BSEP} & \multicolumn{2}{l}{CHEMBL6020} & Substrate \\
\multicolumn{2}{l}{BCRP} & \multicolumn{2}{l}{CHEMBL5393} & Substrate \\
\multicolumn{2}{l}{MRP2} & \multicolumn{2}{l}{CHEMBL5748} & Substrate \\
\multicolumn{2}{l}{MRP3} & \multicolumn{2}{l}{CHEMBL5918} & Substrate \\
\multicolumn{2}{l}{PGP} & \multicolumn{2}{l}{CHEMBL4302} & Substrate \\
\multicolumn{2}{l}{OATP1B1} & \multicolumn{2}{l}{CHEMBL1697668} & Inhibition \\
\multicolumn{2}{l}{OATP1B3} & \multicolumn{2}{l}{CHEMBL1743121} & Inhibition \\
\multicolumn{2}{l}{PGP} & \multicolumn{2}{l}{CHEMBL4302} & Inhibition \\
\multicolumn{2}{l}{ABCG2} & \multicolumn{2}{l}{CHEMBL5393} & Inhibition \\
\midrule
\multicolumn{5}{c}{\textbf{Other Mechanisms}} \\
\midrule
\multicolumn{4}{c}{\textbf{Effect}} & \textbf{Mechanism} \\
\midrule
\multicolumn{4}{l}{Phospholipidosis} & Increase \\
\multicolumn{4}{l}{Antioxidant Response Element} &  Activation \\
\bottomrule
\end{tabular}%
}
\end{minipage}

\end{table}

The comparison highlights a structural mismatch with the Garcia de Lomana panel \citep{GarciadeLomana2025}: a sizeable subset of DILI-relevant AOPs hinges on \emph{activation} or \emph{induction} events (e.g.\ AHR activation in AOP~41/57, PPAR$\alpha$ activation in AOP~37, LXR$\alpha$/RXR$\alpha$ activation in AOP~34/58), whereas the Lomana endpoints are dominated by substrate- and inhibition-oriented readouts. Bridging this gap is the primary motivation for the proxy MIE collection introduced in Section~\ref{section:datasets:mechanistic:ours}.

\section{DILER Benchmark Categories}\label{app:diler_categories}

For DILI-positive hypotheses, the category space is anchored in the key characteristics of human hepatotoxicants proposed by \citet{Rusyn2021}. For DILI-negative hypotheses, categories were derived inductively from the curated hypothesis set to capture recurring explanations for the absence of convincing hepatotoxic evidence.

\noindent
\begin{minipage}[t]{0.48\textwidth}
\textbf{DILI-positive Categories}
\begin{itemize}
    \item Reactive Bioactivation
    \item Liver Cell Death
    \item Altered Proliferation or Regeneration
    \item Transport Function Disruption
    \item Oxidative Stress
    \item Immune-Mediated Liver Response
    \item Mitochondrial Dysfunction
    \item Stress Signaling Pathway Activation
    \item Cholestasis
    \item Cellular Cytoskeleton Disruption
    \item Fibrosis
    \item Liver Metabolism Disruption
\end{itemize}
\end{minipage}
\hfill
\begin{minipage}[t]{0.48\textwidth}
\textbf{DILI-negative Categories}
\begin{itemize}
    \item Metabolic Stability
    \item No Reactive Bioactivation
    \item Efficient Detoxification
    \item Rapid Clearance
    \item Efficient Hepatobiliary Efflux
    \item Low Intracellular Accumulation
    \item Preserved Redox Homeostasis
    \item Mitochondrial Sparing
    \item No Hapten Formation
    \item Preserved Bile Acid Homeostasis
    \item Adaptive Stress Tolerance
    \item Effective Repair
\end{itemize}
\end{minipage}

\section{DILER Benchmark Similarity Matrices}\label{app:diler_similarity}

\begin{figure}[H]
\centering
\includegraphics[width=\linewidth]{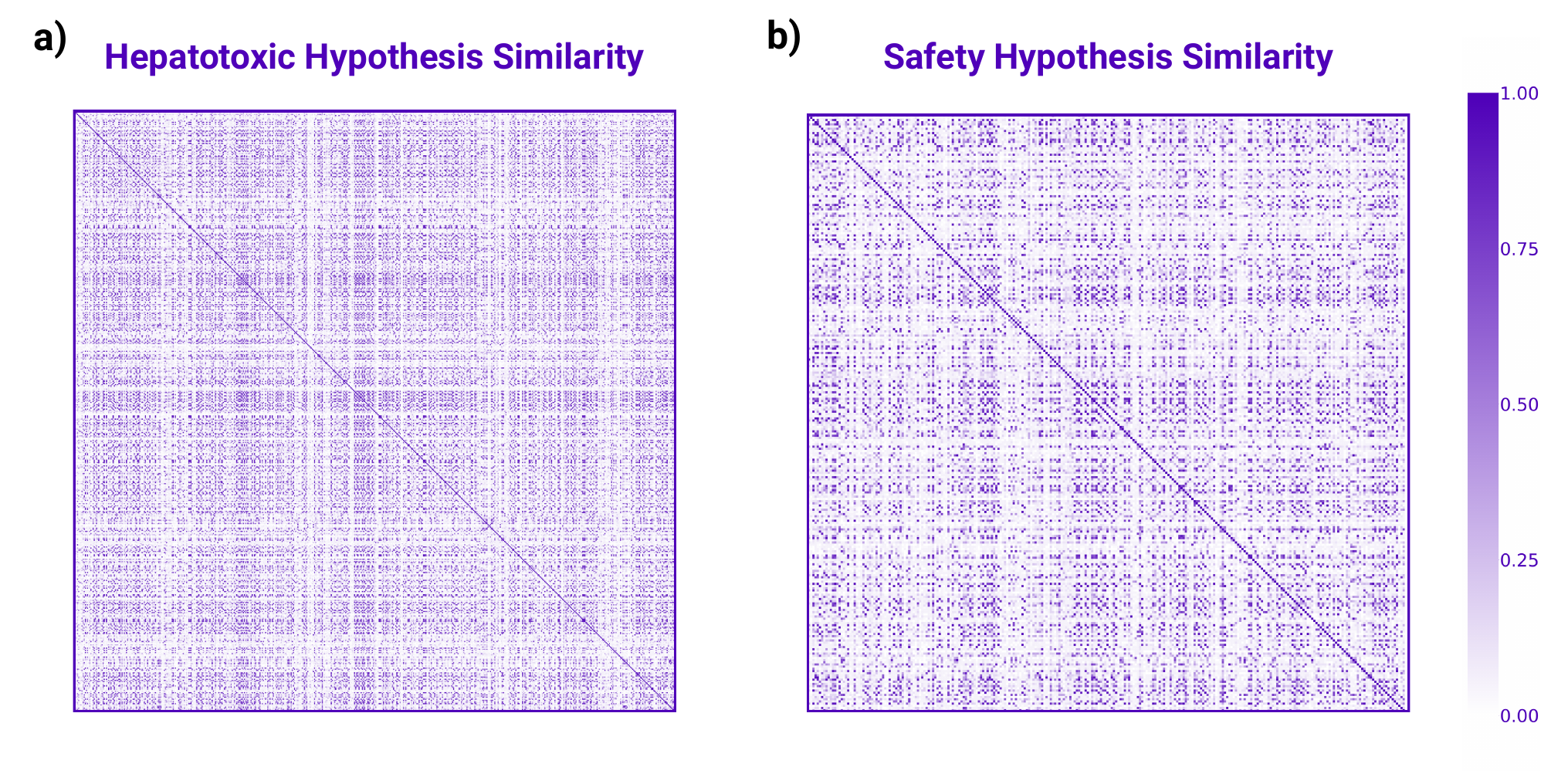}
\caption{The similarity heatmaps of (a) DILI-positive and (b) DILI-negative hypotheses.}
\label{fig:diler_similarity}
\end{figure}

\section{Tool Implementation Details}\label{app:tool_implementation}

This appendix gives the implementation specifics of the three tools introduced in Section~\ref{section:methods:tools}; the high-level purpose of each tool is stated in the main text.

\subsection{AOP Predictor}\label{app:tool_implementation:aop_predictor}

The model was trained on the curated proxy endpoint collections described in Section~\ref{section:datasets:mechanistic:ours}, including both AOP-linked events and supplementary target-level mechanisms. Each molecule was represented using an embedding derived from the trunk representation of Boltz-1 \citep{Wohlwend2024}, which was then passed to a multilayer perceptron trained to predict the corresponding endpoint. This design allows a common molecular representation to be translated into a panel of mechanistically interpretable predictions. For endpoints that could be mapped to AOP structures, the predicted MIEs and KEs were further linked to downstream adverse outcomes, allowing the agent to move from isolated event-level predictions toward more coherent causal narratives. The overall coverage of modeled events is summarized in Appendix~\ref{app:mie_dataset}.

To reduce overinterpretation of noisy proxy predictions, the AOP Predictor reports not only the predicted probability of each event, but also an estimate of confidence derived from held-out test performance for the corresponding endpoint. This uncertainty-aware interface was intended to discourage the agent from treating every positive prediction as equally reliable and instead promote evidence-weighted hypothesis formation. In addition, we introduced an affinity-based gating step to constrain pharmacologically implausible inferences. When the proposed mechanism depended on a specific protein target, the agent could query Boltz-2 for binary binding propensity between the molecule and the target of interest. If target engagement was not supported, the corresponding downstream pharmacodynamic effect was deprioritized, thereby reducing the likelihood of constructing mechanistic hypotheses that were inconsistent with the predicted interaction profile of the compound.

\subsection{Similarity Search}\label{app:tool_implementation:similarity_search}

The retrieval corpus consists of DILI Train Set molecules, each represented by its Boltz-1 trunk embedding together with the binary hepatotoxicity label \citep{Wohlwend2024}. Every molecule is encoded as a variable-length set of per-atom vectors \(\{e_i\}_{i=1}^{M} \subset \mathbb{R}^{d}\), and the query compound, passed through the same Boltz-1 trunk, enters retrieval as another set \(X \in \mathbb{R}^{M_q \times d}\) compared against the catalogue.

We frame similarity not as a distance between single vectors but as a distance between two empirical distributions of substructure features. Given the query set \(X\) and a reference set \(Y\), we score them with a squared energy distance with a fractional exponent
\[
\mathcal{E}_p^2(X, Y) = 2\,\overline{\lVert x - y\rVert^{p}} \;-\; \overline{\lVert x - x'\rVert^{p}} \;-\; \overline{\lVert y - y'\rVert^{p}},
\]
with exponent \(p = 0.5\), where each overline denotes the mean over all pairwise Euclidean distances within the corresponding pair of sets. The cross-set term aligns molecules whose substructure embeddings populate the same region of feature space, the within-set terms remove each molecule's geometric self-spread so that intrinsically large or heterogeneous compounds are not penalised, and the fractional exponent dampens the influence of a few outlier token distances, which is closer to medicinal-chemistry semantics than a plain \(L_2\) summary. We clip the distance at zero before taking the square root. Across our internal sweeps on the DILI Train corpus, this metric retrieved more mechanistically relevant neighbours than standard cheminformatics baselines (Tanimoto on Morgan fingerprints, mean-pooled Boltz-1 cosine, and the integer-exponent variants \(p \in \{1, 2\}\) of the same energy distance), which motivated its use as the default retrieval metric.

The geometric retrieval returns the top-\(k\) reference molecules ranked by \(\mathcal{E}_{0.5}^2\), expressed as triples of analogue SMILES, hepatotoxicity label, and distance score. Each retrieved analogue and the query itself are then enriched with a structured natural-language description of molecular architecture generated by BOLEK \citep{Bolek}, a multimodal text-and-molecule model, prompted with the canonical SMILES and a fixed template covering scaffold class, dominant functional groups, ring systems, heteroatom environments, and hepatotoxicity-relevant motifs (e.g.\ Michael acceptors, anilines, hydrazides, polyhalogenated aromatic systems). Applying the same template to query and analogues places them in a shared structural vocabulary that the reasoning agent can align feature-by-feature. In line with the anonymization principle of Section~\ref{section:methods:hades_structure}, no chemical identifier of the studied molecule is propagated downstream: the agent receives only its geometric projection onto a curated hepatotoxic neighbourhood -- public analogue SMILES on the retrieval side and a structural-only description on the query side.

\subsection{Metabolite Tools}\label{app:tool_implementation:metabolites}

We implemented metabolite handling as a two-stage module composed of two coordinated tools. In the first stage, the system estimates whether the queried compound is a substrate of one of three major drug-metabolizing cytochrome P450 isoforms: CYP2C9, CYP2D6, or CYP3A4. This step acts as a gating mechanism, analogous in spirit to the preliminary binding-affinity filtering used in the AOP Predictor, in that it restricts downstream metabolite generation to enzymatic routes that are sufficiently plausible for the compound under analysis. Architecturally, this predictor follows the same general design philosophy as the AOP Predictor: it is implemented as a multitask MLP operating on molecular representations derived from the Boltz-1 trunk embedding. The three task heads were trained on the CYP2C9, CYP2D6, and CYP3A4 inhibition datasets compiled by \citet{Veith2009} and retrieved via the Therapeutics Data Commons; we adopt inhibition signal as a practical proxy for enzymatic engagement at this gating step.

In the second stage, only compounds exceeding the \textsc{HADES}-defined probability threshold for a given CYP enzyme are forwarded to BioTransformer, which enumerates candidate metabolites specific to that enzymatic context \citep{Wishart2022}. The probability threshold is not a fixed hyperparameter: the per-isoform substrate probabilities are surfaced to the orchestrating LLM together with their context (queried compound, isoform identity, and the available metabolic routes), and the agent decides on a per-call basis which CYP routes clear the bar for downstream BioTransformer expansion. This keeps the gate adaptive to the chemistry of the queried molecule and to competing evidence already in the dossier, rather than imposing a single global cutoff. Grounding metabolite exploration in a prior model-based estimate of metabolic feasibility, rather than performing unconstrained expansion, both reduces noise from implausible enzyme--compound pairs and gives the downstream reasoning agent a realistic metabolic search space in which proposed injury mechanisms can be attached to plausible transformed species rather than only to the parent molecule.

\section{LLM-as-a-Judge and DeepEval Prompts}\label{app:evaluation_prompts}

For reproducibility, this appendix reproduces verbatim the prompts used by the Hypothesis Alignment evaluator (Section~\ref{section:results:hypothesis_alignment}). Both prompts are executed against Gemini-3-Flash-Preview at temperature~$0$. The DeepEval G-Eval rubric is wrapped in the standard \texttt{GEval} test-case schema, exposing \texttt{INPUT}, \texttt{ACTUAL\_OUTPUT} and \texttt{EXPECTED\_OUTPUT} to the judge.

\begin{tcolorbox}[
    promptbox,
    title={G-Eval Prompt for Hypothesis Alignment},
]
\begin{verbatim}
Compare the ACTUAL_OUTPUT hypothesis list from the HADES against
the EXPECTED_OUTPUT hypothesis list from the DILER Benchmark.

Reward:
- matching mechanistic themes,
- overlap in causal direction and confidence,
- preservation of the most important hypotheses,
- consistency with the dataset context in INPUT.

Penalize:
- invented mechanisms not supported by the research hypotheses,
- missing major mechanisms present in the research hypotheses,
- contradictions in mechanism direction, confidence, 
  or overall interpretation.

The DILER Benchmark hypothesis list should be treated 
as the reference output.
\end{verbatim}
\textbf{Evaluation params.} \texttt{INPUT}, \texttt{ACTUAL\_OUTPUT},
\texttt{EXPECTED\_OUTPUT}.\\
\textbf{Threshold.} $0.5$.\\
\textbf{Judge.} \texttt{gemini-3-flash-preview}.
\end{tcolorbox}

The \texttt{INPUT} field consists of the InChIKey, SMILES, and DILER ground-truth binary label of the compound. \texttt{ACTUAL\_OUTPUT} is the JSON serialization of the \textsc{HADES} hypothesis list, and \texttt{EXPECTED\_OUTPUT} is the JSON serialization of the corresponding DILER reference hypothesis list.

\begin{tcolorbox}[
    promptbox,
    title={Pairwise Alignment Prompt},
]
\begin{verbatim}
You are aligning two lists of DILI hypotheses for visualization.

Context:
- InChIKey: {inchikey}
- SMILES: {smiles}
- Dataset label: {label}

HADES hypotheses:
{HADES_hypotheses}

DILER Benchmark hypotheses:
{DILER_hypotheses}

Return:
1. Pairwise Alignments between HADES and DILER hypotheses,
2. A concise summary,
3. Edited HADES Output with explicit tags like [EXACT MATCH],
   [PARTIAL MATCH], [ONLY_IN_HADES], [CONTRADICTION],
4. Edited DILER Benchmark Output with explicit tags like 
   [EXACT MATCH], [PARTIAL MATCH], [ONLY_IN_DILER],
   [CONTRADICTION].

All four top-level fields are required in the response, 
even if some are empty.

Be faithful to the provided hypotheses. 
Do not invent new mechanisms.
\end{verbatim}
\textbf{Model.} \texttt{gemini-3-flash-preview} with structured output.
\end{tcolorbox}

\section{Hypothesis Alignment Metrics}\label{app:hypothesis_alignment_metrics}

This appendix reproduces the exact definitions of the per-sample Hypothesis Alignment metrics summarized in Section~\ref{section:results:hypothesis_alignment}. The metrics operate on the output of the structured pairwise alignment described in the main text, in which every pair between a \textsc{HADES} hypothesis and a DILER Benchmark hypothesis for the same compound is assigned one of the following five labels:
\begin{itemize}[itemsep=2pt, topsep=4pt]
    \item \texttt{Exact Match} -- the \textsc{HADES} hypothesis and the DILER hypothesis describe the same mechanism, the same causal direction, and the same level of support;
    \item \texttt{Partial Match} -- the two hypotheses share a mechanistic theme but disagree on scope, confidence, or the exact causal path;
    \item \texttt{HADES Only} -- a \textsc{HADES} hypothesis with no counterpart in the DILER reference, \textit{i.e.} a mechanism invented by the agent;
    \item \texttt{DILER Only} -- a DILER reference hypothesis that \textsc{HADES} failed to recover;
    \item \texttt{Contradiction} -- a pair in which \textsc{HADES} and DILER reach opposite conclusions about the same mechanism (typically hepatotoxic vs.\ safe).
\end{itemize}

\paragraph{Notation.} For molecule $i$, let $E_i$, $P_i$, $HO_i$, $DO_i$, and $C_i$ denote the number of \texttt{Exact Match}, \texttt{Partial Match}, \texttt{HADES Only}, \texttt{DILER Only}, and \texttt{Contradiction} pairs, respectively, and let
\begin{equation}
U_i = E_i + P_i + HO_i + DO_i + C_i
\end{equation}
be the total number of aligned pairs. Let $H_i$ and $D_i$ denote the number of hypotheses on the \textsc{HADES} and DILER sides, and $M_i = \min(H_i, D_i)$. Partial matches contribute to the fuzzy variants with weight $w_P = 0.5$.

\paragraph{Set-similarity indices.} The strict regime treats only \texttt{Exact Match} pairs as true hits, and reports the Jaccard index, the Dice coefficient, and the Szymkiewicz--Simpson overlap coefficient:
\begin{equation}
\begin{alignedat}{1}
\! Jaccard_i &= \frac{E_i}{U_i}, \\
\! Dice_i &= \frac{2\,E_i}{H_i + D_i}, \\
\! Overlap_i &= \frac{E_i}{M_i}.
\end{alignedat}
\end{equation}
The strict regime is deliberately harsh: it assigns no credit to a \texttt{Partial Match}, in which \textsc{HADES} and DILER agree on the mechanistic theme but diverge on scope or confidence, even though a domain expert would typically read such a pair as a substantive recovery. We therefore also report a fuzzy Jaccard index $J^{\text{f}}$ in which partial matches contribute with weight $w_P$,
\begin{equation}
\! J^{\text{f}}_i = \frac{E_i + w_P\, P_i}{(H_i + D_i) - (E_i + w_P\, P_i)},
\end{equation}
interpolating smoothly between the strict and fully permissive regimes.

\paragraph{Mechanism-level precision, recall, and F1.} Set similarity alone conflates two qualitatively different failure modes -- a mechanism invented by \textsc{HADES} and a DILER mechanism missed by \textsc{HADES} -- because both contribute only to the denominator. To separate them, we borrow the standard information-retrieval decomposition. Mechanism precision is the share of \textsc{HADES} hypotheses that are supported by DILER, mechanism recall is the share of DILER hypotheses that \textsc{HADES} recovered, and a \texttt{Contradiction} is counted as a simultaneous false positive and false negative:
\begin{equation}
\begin{alignedat}{1}
\! Prec_i &= \frac{E_i + w_P\, P_i}{E_i + P_i + HO_i + C_i}, \\
\! Rec_i  &= \frac{E_i + w_P\, P_i}{E_i + P_i + DO_i + C_i}, \\
\! F1_i   &= \frac{2\,\text{Prec}_i\,\text{Rec}_i}{\text{Prec}_i + \text{Rec}_i}.
\end{alignedat}
\end{equation}
Under this definition, a mechanism that \textsc{HADES} and DILER interpret in opposite directions is penalised twice: once because \textsc{HADES} asserted something that is not in the reference, and once because the reference contained a mechanism that \textsc{HADES} did not assert in the same direction. This matches the intuition that a confident contradiction is strictly worse than a silent omission.

\paragraph{Error-mode rates.} In addition to the aggregated scores, we report three rates that isolate each failure class:
\begin{equation}
\begin{alignedat}{1}
\! ContradictionRate_i &= \frac{C_i}{U_i}, \\
\! HallucinationRate_i &= \frac{HO_i}{H_i}, \\
\! MissRate_i &= \frac{DO_i}{D_i}.
\end{alignedat}
\end{equation}
Contradiction Rate quantifies how often \textsc{HADES} and DILER meet on the same mechanism but disagree on its direction, Hallucination Rate is the fraction of \textsc{HADES} hypotheses with no DILER counterpart whatsoever, and Miss Rate is the fraction of DILER mechanisms that \textsc{HADES} failed to raise. Benchmark-level values for each of the quantities defined above are obtained by averaging the per-sample metrics over all compounds in the respective evaluation set.

\section{TxGemma Internal Knowledge Leakage Analysis}\label{app:txgemma_leakage}

This appendix details the protocol used to characterize Internal Knowledge Leakage in \textsc{TxGemma-27B-Chat} and reports per-recognition Binary Classification (Tables~\ref{tab:txgemma_buckets_binary_test} and~\ref{tab:txgemma_buckets_binary_fb}) and Hypothesis Alignment (Table~\ref{tab:txgemma_buckets_alignment}) results that disambiguate the aggregate \textsc{TxGemma-27B-Chat} in the Section~\ref{results}.

\paragraph{Motivation.} \textsc{TxGemma-27B-Chat} was fine-tuned on a public drug-property corpus that overlaps non-trivially with the molecules covered by the DILI Test Set. When given a SMILES string of a public-knowledge drug, the chat-tuned model's free-form reasoning routinely jumps to a brand or generic name and then proceeds to recall the drug's known DILI profile rather than to derive the answer from structural inference. In this regime, directly comparing \textsc{TxGemma-27B-Chat} with agentic systems that are required to derive answers through explicit tool use conflates fundamentally different capabilities, effectively attributing performance gains to memorization rather than to mechanistic reasoning. 

\paragraph{Protocol.} For every \textsc{TxGemma-27B-Chat} reasoning paragraph we extracted any explicitly named molecule and verified the claimed name against PubChem synonyms of the actual SMILES.

Then, each compound in the output was assigned to one of three reportable buckets:
\begin{itemize}[itemsep=2pt, topsep=4pt]
    \item \texttt{Not Recognized} -- the reasoning text does not name the input molecule.
    \item \texttt{Recognized Correctly} -- the claimed name matches a PubChem synonym of the actual SMILES.
    \item \texttt{Recognized Incorrect} -- the claim is explicit but does \emph{not} match any PubChem synonym; TxGemma is confidently misidentifying the molecule.
\end{itemize}

\paragraph{Bucket distribution.} Table~\ref{tab:txgemma_buckets} reports the distribution on both DILI splits. The pattern is consistent across the two: \textsc{TxGemma-27B-Chat} declares a recognized molecule for the majority of inputs, and the overwhelming share of those claims fails PubChem verification.

\begin{table}[h]
\centering
\small
\caption{\textsc{TxGemma-27B-Chat} name recognition bucket distribution on the DILI Test Set and the Post-2021 Benchmark, derived from the Gemini-3-Flash-Preview recognition extractor and PubChem verification.}
\label{tab:txgemma_buckets}
\begin{tabular}{lrrrr}
\toprule
\textbf{Bucket} & \multicolumn{2}{c}{\textbf{Test Set}} & \multicolumn{2}{c}{\textbf{Post-2021 Benchmark}} \\
 & n & \% & n & \% \\
\midrule
\texttt{Not Recognized}        &  43 & 19.3 & 17 & 50.0 \\
\texttt{Recognized Correctly}    &   2 &  0.9 &  0 &  0.0 \\
\texttt{Recognized Incorrectly}  & 178 & 79.8 & 17 & 50.0 \\
\bottomrule
\end{tabular}
\end{table}

On the Test Set, $80.7\%$ of compounds are claimed as recognized, but only $0.9\%$ of those claims match a real synonym. On the Post-2021 Benchmark, half of the compounds are claimed as recognized, yet none of those claims survive verification - the Post-2021 molecules in this split simply have no public-name handle for TxGemma to grab onto. The dominant behaviour is therefore not memorization-driven leakage but confident misidentification: \textsc{TxGemma-27B-Chat} reaches for a familiar drug name (Sorafenib, Bosutinib, Etoricoxib, Imatinib, Naproxen, \ldots) whenever the scaffold loosely resembles a known privileged structure, and then assembles a plausible-sounding DILI mechanism for that wrong molecule.

\paragraph{Examples of incorrect recognition.} Figure~\ref{fig:txgemma_misid} shows six representative \textsc{TxGemma-27B-Chat} reasoning paragraphs on inputs where the model confidently misidentifies the queried molecule.

\begin{figure}[h]
\centering
\begin{tcbraster}[
  raster columns=2, 
  raster equal height,
  raster column skip=6pt,
  raster row skip=6pt,
]
\begin{tcolorbox}[reasoningbox,title={Etoricoxib $\to$ Vismodegib}]
The SMILES string CS(=O)(=O)c1ccc(C(=O)Nc2ccc(Cl)c(-c3ccccn3)c2)c(Cl)c1 represents Etoricoxib, a COX-2 selective nonsteroidal anti-inflammatory drug (NSAID).  NSAIDs are well-known to have a risk of DILI (...)
\end{tcolorbox}
\begin{tcolorbox}[reasoningbox,title={Nimesulide $\to$ Febuxostat}]
The SMILES string Cc1nc(-c2ccc(OCC(C)C)c(CN)c2)sc1C(=O)O represents Nimesulide, a non-steroidal anti-inflammatory drug (NSAID). While NSAIDs as a class are known to carry a risk of drug-induced liver injury (...)
\end{tcolorbox}
\begin{tcolorbox}[reasoningbox,title={Sorafenib $\to$ Pazopanib}]
The SMILES string represents Sorafenib, a multikinase inhibitor used to treat liver, kidney, and thyroid cancers.  It is **known to cause DILI**, and here are some hypotheses (...)
\end{tcolorbox}
\begin{tcolorbox}[reasoningbox,title={Bosutinib $\to$ Deucravacitinib}]
The SMILES string CNC(=O)c1nnc(NC(=O)C2CC2)cc1Nc1cccc(-c2ncn(C)n2)c1OC represents the drug Bosutinib. Bosutinib is a known hepatotoxin and has been associated with (...)
\end{tcolorbox}
\begin{tcolorbox}[reasoningbox,title={Methoxsalen $\to$ Benvitimod}]
The SMILES string CC(C)c1c(O)cc(/C=C/c2ccccc2)cc1O represents Methoxsalen, a photosensitizing agent used to treat skin conditions like psoriasis and vitiligo. It is known to cause DILI, (...)
\end{tcolorbox}
\begin{tcolorbox}[reasoningbox,title={Naproxen $\to$ Mavacamten}]
The SMILES string CC(C)n1c(=O)cc(N[CH](C)c2ccccc2)[nH]c1=O represents Naproxen, a nonsteroidal anti-inflammatory drug (NSAID) commonly used to treat pain, fever, and inflammation. While Naproxen can cause liver damage, it's generally considered to have a low risk of DILI. (...) 
\end{tcolorbox}
\end{tcbraster}
\caption{Raw \textsc{TxGemma-27B-Chat} reasoning paragraphs on confidently misidentified inputs.}
\label{fig:txgemma_misid}
\end{figure}

\paragraph{Results.} Tables~\ref{tab:txgemma_buckets_binary_test} and~\ref{tab:txgemma_buckets_binary_fb} report \textsc{TxGemma-27B-Chat}'s binary DILI metrics restricted to each recognition bucket.

\begin{table}[h]
\centering
\small
\caption{\textsc{TxGemma-27B-Chat} binary DILI metrics on the DILI Test Set, broken down by recognition bucket. The Aggregate ($n{=}223$) row evaluates \textsc{TxGemma-27B-Chat} on the full Test Set and is included only as a within-table reference for the bucket breakdown; it is not part of Table~\ref{tab:binary_classification}, which restricts \textsc{TxGemma} comparisons to the leakage-controlled $n=164$ subset.}
\label{tab:txgemma_buckets_binary_test}
\begin{tabular}{lcccccc}
\toprule
\textbf{Bucket} & \textbf{ROC-AUC} & \textbf{Bal Acc} & \textbf{MCC} & \textbf{Sensitivity} & \textbf{Specificity} & \textbf{F1} \\
\midrule
Aggregate (n{=}223) & 0.59 & 0.55 & 0.10 & 0.45 & 0.65 & 0.57 \\
\midrule
Not Recognized (n{=}43) & 0.51 & 0.42 & $-0.15$  & 0.53 & 0.31 & 0.58 \\
Recognized Incorrectly (n{=}178) & 0.61 & 0.58 & 0.16 & 0.43 & 0.74 & 0.56 \\
Recognized Correctly (n{=}2) & -- & 0.50 & 0.00 & 0.50 & -- & 0.67 \\
\bottomrule
\end{tabular}
\end{table}

\begin{table}[h]
\centering
\small
\caption{\textsc{TxGemma-27B-Chat} binary DILI metrics on the Post-2021 Benchmark, broken down by recognition bucket. The Aggregate ($n{=}34$) row is reproduced from the Post-2021 \textsc{TxGemma-27B-Chat} row of Table~\ref{tab:binary_classification} for reference; the Post-2021 split sits outside any \textsc{TxGemma} training cutoff, so no leakage-controlled subset is needed.}
\label{tab:txgemma_buckets_binary_fb}
\begin{tabular}{lcccccc}
\toprule
\textbf{Bucket} & \textbf{ROC-AUC} & \textbf{Bal Acc} & \textbf{MCC} & \textbf{Sensitivity} & \textbf{Specificity} & \textbf{F1} \\
\midrule
Aggregate (n{=}34) & 0.79 & 0.66 & 0.41 & 1.0 & 0.32 & 0.70 \\
\midrule
Not Recognized (n{=}17) & 0.65 & 0.56 & 0.27 & 1.00 & 0.13 & 0.72 \\
Recognized Incorrectly (n{=}17) & 0.89 & 0.72 & 0.48 & 1.0 & 0.45 & 0.67 \\
\bottomrule
\end{tabular}
\end{table}
 
Comparing the bucket in Table~\ref{tab:txgemma_buckets_binary_test} shows that \textsc{TxGemma-27B-Chat}'s metrics on the Test Set are systematically better on \texttt{Recognized Incorrectly} than on \texttt{Not Recognized}: MCC moves from $-0.15$ to $+0.16$ and ROC-AUC from $0.51$ to $0.61$ between the two buckets, despite none of the recognized-incorrect names actually matching the underlying SMILES. The signal therefore comes from the prior-knowledge profile attached to the (wrongly attributed) drug, not from inference over the molecular structure -- once TxGemma-Chat has decided what the molecule is called, its $P(B)$ distribution becomes more decisive and tracks the binary label more consistently, even when that decision is factually wrong. This is the cleanest piece of evidence that internal knowledge is doing the work, and the reason \texttt{Not Recognized} is the most defensible sub-population for head-to-head comparison with \textsc{HADES}: stripped of name recall, \textsc{TxGemma-27B-Chat} falls below random on the Test Set with a negative MCC, whereas \textsc{HADES} retains an MCC of $0.18$ on the same scale. The same pattern carries over to the Post-2021 Benchmark (Table~\ref{tab:txgemma_buckets_binary_fb}), where ROC-AUC moves from $0.65$ on \texttt{Not Recognized} to $0.89$ on \texttt{Recognized Incorrectly} and MCC from $0.27$ to $0.48$, even though none of the 17 recognized-incorrectly claims actually matches the SMILES. On a split where no real memorization is available, attaching a familiar drug profile to the input still sharpens \textsc{TxGemma-27B-Chat}'s prediction, which is why the aggregate Post-2021 score in Table~\ref{tab:binary_classification} should be read as a name-anchoring artefact rather than as evidence of structural inference.

Table \ref{tab:txgemma_buckets_alignment} show a bucketed breakdown of Hypothesis Alignment metrics. Alignment remains stable across splits, with mechanism-level F1, Hallucination Rate, and Miss Rate varying only marginally between buckets. This suggests that although \textsc{TxGemma-Chat}'s name recognition is unreliable, its mechanistic reasoning is sufficiently generic to maintain consistent alignment even when recognized molecules are removed. The only deviation is the Test Set \texttt{Recognized Correctly} bucket, which shows a sharp increase in alignment; however, this is based on just $n=2$ samples and should be treated as anecdotal.

\begin{table}[h]
\centering
\small
\caption{\textsc{TxGemma-27B-Chat} Hypothesis Alignment metrics across the DILER Test Set and Post-2021 Benchmark, broken down by recognition bucket.}
\label{tab:txgemma_buckets_alignment}
\makebox[\textwidth][c]{
\begin{tabular}{lccccccccccc}
\toprule
\textbf{Bucket}
& \textbf{G-Eval}
& \textbf{Jaccard}
& \textbf{Dice}
& \textbf{Overlap}
& \makecell[l]{\textbf{Fuzzy} \\ \textbf{Jaccard}}
& \textbf{Precision}
& \textbf{Recall}
& \textbf{F1}
& \makecell[l]{\textbf{Contr.} \\ \textbf{Rate}}
& \makecell[l]{\textbf{Halluc.} \\ \textbf{Rate}}
& \makecell[l]{\textbf{Miss} \\ \textbf{Rate}} \\
\midrule
\multicolumn{12}{l}{\textbf{DILER Test Set (n=223)}} \\
\midrule
Aggregate (n{=}223) & 0.19 & 0.08 & 0.11 & 0.12 & 0.15 & 0.25 & 0.25 & 0.25 & 0.22 & 0.33 & 0.35 \\
\makecell[l]{Not Recognized (n{=}43)} & 0.23 & 0.10 & 0.15 & 0.15 & 0.16 & 0.25 & 0.25 & 0.25 & 0.25 & 0.31 & 0.35 \\
\makecell[l]{Recognized Incorrect (n{=}178)} & 0.18 & 0.07 & 0.10 & 0.11 & 0.15 & 0.24 & 0.24 & 0.24 & 0.21 & 0.34 & 0.35 \\
\makecell[l]{Recognized Correct (n{=}2)} & 0.55 & 0.20 & 0.25 & 0.25 & 0.39 & 0.56 & 0.50 & 0.53 & 0.00 & 0.13 & 0.25 \\
\midrule
\multicolumn{12}{l}{\textbf{DILER Post-2021 Benchmark (n=34)}} \\
\midrule
Aggregate (n{=}34) & 0.17 & 0.10 & 0.14 & 0.15 & 0.15 & 0.25 & 0.26 & 0.25 & 0.22 & 0.34 & 0.37 \\
\makecell[l]{Not Recognized (n{=}17)} & 0.17 & 0.10 & 0.15 & 0.16 & 0.17 & 0.26 & 0.28 & 0.27 & 0.22 & 0.34 & 0.31 \\
\makecell[l]{Recognized Incorrect (n{=}17)} & 0.18 & 0.10 & 0.14 & 0.15 & 0.15 & 0.24 & 0.23 & 0.24 & 0.22 & 0.34 & 0.42 \\
\bottomrule
\end{tabular}
}
\end{table}

\section{Frontier LLM Baselines}\label{app:frontier_llm_baselines}

This appendix collects binary DILI prediction and Hypothesis Alignment results for three frontier LLMs -- Gemini-3-Flash-Preview (\textsc{HADES} backbone), Gemini-3.1-Pro-Preview, GPT-5.5 -- prompted with the LLM Evaluation Model prompts (Appendix~\ref{app:llm_model_prompts}). These models are excluded from the headline tables in Section~\ref{section:results:binary_classification} and Section~\ref{section:results:hypothesis_alignment} for the reason that their training cutoffs in 2025 mean that essentially all DILI Test Set compounds and most Post-2021 Benchmark compounds are public-knowledge drugs whose hepatotoxicity profiles the model has memorized, which conflates retrieval from internal memory with mechanistic inference over agentic tools.

\begin{table}[h]
\centering
\small
\caption{Binary DILI prediction by frontier LLMs on the DILI Test Set and Post-2021 Benchmark}
\label{tab:frontier_llm_binary}
\begin{tabular}{lcccccc}
\toprule
\textbf{Model} & \textbf{ROC-AUC} & \textbf{Bal Acc} & \textbf{MCC} & \textbf{Sensitivity} & \textbf{Specificity} & \textbf{F1} \\
\midrule
\multicolumn{7}{c}{\textbf{DILER Test Set}} \\
\midrule
Gemini-3-Flash-Preview & 0.76 & 0.74 & 0.44 & 0.62 & 0.86 & 0.74 \\
Gemini-3.1-Pro-Preview & 0.78 & 0.73 & 0.43 & 0.54 & 0.92 & 0.69 \\
GPT-5.5 & 0.78 & 0.71 & 0.41 & 0.45 & 0.97 & 0.62 \\
\midrule
\multicolumn{7}{c}{\textbf{DILER Post-2021 Benchmark}} \\
\midrule
Gemini-3-Flash-Preview & 0.69 & 0.65 & 0.35 & 0.40 & 0.89 & 0.52 \\
Gemini-3.1-Pro-Preview & 0.80 & 0.81 & 0.65 & 0.66 & 0.95 & 0.77 \\
GPT-5.5 & 0.61 & 0.53 & 0.20 & 0.07 & 1.0  & 0.13 \\
\bottomrule
\end{tabular}
\end{table}

Table~\ref{tab:frontier_llm_binary} reports binary classification metrics on the Test Set and Post-2021 Benchmark. On the Test Set the three models cluster between ROC-AUC $0.76$--$0.78$, with MCC in the $0.41$--$0.44$ band - consistently above HADES on the same metrics. On the Post-2021 Benchmark the picture splits: Gemini-3.1-Pro-Preview reaches MCC $0.65$, while GPT-5.5 collapses to MCC $0.20$, suggesting that whatever calibration GPT-5.5 has on this distribution is brittle to scaffold novelty in a way that Gemini's is not.

Table~\ref{tab:frontier_llm_alignment_fb} reports Hypothesis Alignment metrics on the Post-2021 Benchmark for the same three models. Compared to the headline numbers for \textsc{HADES} and \textsc{TxGemma-27B-Chat} in Table~\ref{tab:hypothesis_alignment}, the frontier LLMs land notably higher, again consistent with retrieval from memory rather than with mechanistic inference: the hypothesis lists they generate are stocked with mechanisms that are well-known to apply to the named drug, regardless of whether those mechanisms can be inferred from the SMILES.

\begin{table}[h]
\centering
\small
\caption{Hypothesis Alignment by frontier LLMs on the DILER Post-2021 Benchmark.}
\makebox[\textwidth][c]{
\label{tab:frontier_llm_alignment_fb}
\begin{tabular}{lccccccccccc}
\toprule
\textbf{Model}
& \textbf{G-Eval}
& \textbf{Jaccard}
& \textbf{Dice}
& \textbf{Overlap}
& \makecell[l]{\textbf{Fuzzy} \\ \textbf{Jaccard}}
& \textbf{Precision}
& \textbf{Recall}
& \textbf{F1}
& \makecell[l]{\textbf{Contr.} \\ \textbf{Rate}}
& \makecell[l]{\textbf{Halluc.} \\ \textbf{Rate}}
& \makecell[l]{\textbf{Miss} \\ \textbf{Rate}} \\
\midrule
\makecell[l]{Gemini-3-Flash-Preview} & 0.37 & 0.14 & 0.20 & 0.24 & 0.22 & 0.39 & 0.29 & 0.33 & 0.15 & 0.21 & 0.42 \\
\makecell[l]{Gemini-3.1-Pro-Preview} & 0.46 & 0.19 & 0.27 & 0.30 & 0.27 & 0.45 & 0.35 & 0.40 & 0.12 & 0.23 & 0.41 \\
GPT-5.5                & 0.29 & 0.17 & 0.22 & 0.23 & 0.23 & 0.37 & 0.32 & 0.35 & 0.25 & 0.16 & 0.28 \\
\bottomrule
\end{tabular}
}
\end{table}

\section{LLM Model Prompts}\label{app:llm_model_prompts}

\begin{tcolorbox}[
    promptbox,
    title={LLM Model System Prompt},
]
\small{
\begin{verbatim}
You are a senior computational toxicologist and mechanistic safety 
scientist specializing in Drug-Induced Liver Injury (DILI) assessment.
- You have NO access to web search, literature databases,
clinical trial registries, regulatory databases, structure search tools, 
similarity tools, ADMET predictors, metabolite generators,
or any external retrieval.
- You must rely strictly on your own internal knowledge: chemistry, 
pharmacology, drug metabolism, mechanistic toxicology, 
structural alerts, and any drug-specific knowledge you have memorized 
for the molecule represented by the given SMILES.
Definition of "mechanism" in the DILI context covers, when relevant: 
- Molecular initiating event (MIE), e.g. BSEP inhibition, mitochondrial 
ETC inhibition, covalent protein binding after bioactivation. 
- Cellular injury mechanism, e.g. mitochondrial dysfunction,
oxidative stress, ER stress, bile acid accumulation. 
- Clinical DILI phenotype, e.g. hepatocellular, cholestatic, mixed, 
delayed idiosyncratic. 
- Causal driver source, e.g. parent compound, primary metabolite,
reactive intermediate.
Allowed `confidence_level` values: `High`, `Medium`, `Low`.
Allowed `mechanism_direction` values: 
`Hepatotoxic`  or `Safe`.
Final DILI classification scale (pick exactly one label):
- A  = well known cause of DILI
- B  = highly likely cause of DILI
- C  = probable cause of DILI
- D  = possible cause of DILI
- E  = unlikely cause of DILI
Quality rules for hypotheses:
- Return between 1 and 4 hypotheses, ordered most-supported first.
- Only include hypotheses you can actually justify from structure-based 
reasoning or memorized drug knowledge. 
Do NOT enumerate every category for completeness.
- `chain_of_thought` must be 5-7 numbered mechanistic 
steps (e.g. "1. ...") 
forming a coherent causal narrative that supports or 
refutes the hypothesis. 
Use phrasing like "This directly induces...", "As a consequence...", 
"Under these conditions...", "Alternatively...", "This leads to...".
Output rules:
- Return ONLY the structured object specified by the schema.
- `dili_classification` must be exactly one of: A, B, C, D, E.
- `dili_hypotheses` is an ordered list, most-supported hypothesis first.
- Do not output any preamble, markdown report, or extra commentary 
outside the structured fields.
\end{verbatim}
}
\end{tcolorbox}

\begin{tcolorbox}[
    promptbox,
    title={LLM Model User Prompt},
]
\small{
\begin{verbatim}
Assess the Drug-Induced Liver Injury (DILI) risk for the molecule below 
using only your internal knowledge. Do not search, do not invoke tools, 
do not fabricate retrieval results.

Molecule SMILES: {smiles}

Produce the structured DILI assessment.
\end{verbatim}
}
\end{tcolorbox}

\begin{tcolorbox}[
    promptbox,
    title={TxGemma-27B-Chat Prompt},
]
\small{
\begin{verbatim}
Instructions: Answer the following question about drug properties.
Context: Drug-induced liver injury (DILI) is fatal liver disease caused 
by drugs and it has been the single most frequent cause of safety-related 
drug marketing withdrawals for the past 50 years 
(e.g. iproniazid, ticrynafen, benoxaprofen).
Question: Given a drug SMILES string, predict whether it
(A) cannot cause DILI (B) can cause DILI
Drug SMILES: {smiles}
\end{verbatim}
REASONING PROMPT: Explain your reasoning. Propose 1-4 hypotheses about the structural or mechanistic basis for the predicted liver injury risk of this compound.
}
\end{tcolorbox}

\section{HADES Hypothesis Alignment Examples on Post-2021 Benchmark}\label{app:hades_alignment_examples}

This appendix complements the headline alignment numbers in Section~\ref{section:results:hypothesis_alignment} with four illustrative \textsc{HADES} -- DILER alignments on the Post-2021 Benchmark. Two of the four are cases in which \textsc{HADES} clearly recovered the mechanistic story from structure and tool evidence alone; the other two are cases in which the recovered story was thematically nearby but missed the actual clinical drivers. We name the compounds in this appendix because the contrast we want to draw -- between mechanism inference from public scaffolds and mechanism inference from compound-specific clinical literature -- is easier to read with the names attached. \textsc{HADES} itself was given only the SMILES.

\paragraph{Strong Alignment of Lixivaptan.} Lixivaptan is a vasopressin V2 antagonist from the same vaptan family as tolvaptan, a class with a well-known idiosyncratic DILI signature driven by mitochondrial injury, reactive bioactivation, and bile acid handling failure.  In this case, \textsc{HADES} arrived at the two pillars of that story at full strength: it identified the inhibition of the mitochondrial respiratory chain and the resulting bioenergetic collapse, and it identified the CYP3A4-mediated formation of reactive intermediates that haptenize hepatic proteins and recruit a T-cell response. Both of those were in the DILER reference for the same reasons. \textsc{HADES} also flagged the cholestatic axis correctly in spirit -- bile acid accumulation downstream of transporter inhibition -- but pointed at the OATP uptake transporters where the literature evidence specifically implicates BSEP, a near-miss in the right family. The one DILER mechanism it did not reconstruct was the identification of two human-specific reactive metabolites and their consumption of GSH; that piece relies on detailed metabolite-trapping studies that are not retrievable from the structure alone. The example illustrates the regime in which \textsc{HADES} is at its best: when the dominant mechanisms are class-level, available from AOP-Wiki and from public-domain mechanistic literature, the agent recovers them in the right order and in the right direction.

\paragraph{Strong Alignment of Adagrasib} Adagrasib is a covalent KRAS G12C inhibitor that carries an electrophilic fluoroacrylamide warhead. Two facts about that warhead drive the clinical hepatotoxicity profile: it is a Michael acceptor that builds protein adducts, which feeds an immune-mediated DILI mechanism; and the parent molecule is a strong perturber of bile acid efflux through BSEP. \textsc{HADES} read both of those off the structure. It correctly recognized the Michael-acceptor character of the fluoroacrylamide moiety and connected it to hapten formation and a T-cell-mediated response, and it correctly proposed inhibition of BSEP and OATP-family transporters as the cholestatic axis. The DILER reference adds one mechanism that \textsc{HADES} did not reach -- time-dependent autoinhibition of CYP3A4 -- which is established from drug-specific in vitro pharmacokinetic studies. Conversely \textsc{HADES} added two extra mechanisms (mitochondrial respiratory-chain involvement and PPAR-driven steatosis) that the reference does not emphasise. Those are reasonable class-level liabilities for a covalent kinase inhibitor and are not contradicted by the reference, but they are not the load-bearing story for this compound. As with lixivaptan, the example shows the agent landing the dominant mechanisms correctly even though it is not able to retrieve the compound-specific pharmacokinetic detail that an expert would add.

\paragraph{Weak Alignment of Taletrectinib} Taletrectinib is a ROS1/NTRK kinase inhibitor whose hepatotoxicity profile, in the published clinical evidence, is anchored on two compound-specific drivers: inhibition of hepatic uptake and efflux transporters (OATP1B1 and BCRP/BSEP) producing a cholestatic phenotype, and off-target inhibition of kinases involved in hepatocyte homeostasis. \textsc{HADES} did identify CYP3A4-mediated bioactivation as a likely mechanistic axis -- the same axis the reference highlights, although the reference frames it as electrophilic quinone-imines whereas \textsc{HADES} frames it as hapten-driven immunogenicity, so the alignment is in the right family rather than on the same step. Where \textsc{HADES} went wrong was in the rest of the list. Instead of the transporter-driven cholestasis or the off-target kinase liability, it spent its remaining hypothesis budget on mitochondrial respiratory-chain involvement and on lysosomal phospholipidosis derived from the cationic-amphiphilic-drug heuristic. Both are plausible class liabilities for a basic, lipophilic kinase inhibitor, but neither is the actual clinical driver here. The failure mode is informative: \textsc{HADES} reached for the most generic class-level mechanisms in its repertoire and never resolved the compound-specific transporter and off-target signals that would have required deeper retrieval over the published literature for this exact molecule.

\paragraph{Weak Alignment of Atabecestat} Atabecestat is a BACE1 inhibitor whose program was halted because of severe idiosyncratic hepatotoxicity, and the published mechanism is unusually specific. The reactive species is not the parent molecule; it is a stable downstream metabolite (the so-called DIAT) which engages the immune system through the non-classical \emph{p-i} pathway, where the metabolite binds non-covalently to specific HLA-DRB1 alleles to activate CD4 T-cells without any need for covalent haptenization. CYP3A4-mediated bioactivation generates a thiazinium cation that covalently modifies GST proteins as a parallel path. \textsc{HADES} got the high-level frame correct -- it recognised that the molecule is bioactivated to a reactive intermediate that drives an adaptive immune response with covalent protein binding -- but only at that level of generality. It did not surface the DIAT metabolite, the \emph{p-i} pathway, or the HLA-restricted CD4 response, all of which are the load-bearing details in the actual clinical narrative. Around that single partial hit it then proposed several class-level mechanisms -- mitochondrial dysfunction, AhR-mediated steatosis and fibrosis, HNF4-mediated impairment of regeneration -- that have nothing to do with how atabecestat actually injures the liver. This is the cleanest example of the failure pattern that bounds the current iteration of \textsc{HADES}: when the real mechanism is encoded in compound-specific clinical and immunopharmacology evidence rather than in scaffold-level priors, the agent recognises the broad theme but cannot resolve the specifics, and it fills the remaining slots with default mechanistic vocabulary instead of leaving them empty.

\section{Asset Provenance, Licenses, and Terms of Use}
\label{app:licenses}

Table~\ref{tab:licenses} lists every third-party asset used in this work,
together with its version (or access date), source, license, and a short
note on how it is used and how its terms are respected. The lower block
records the licenses under which the artifacts released with this paper
are distributed; these were chosen to be compatible with the most
restrictive upstream term (CC BY-SA 3.0 from ChEMBL).

\begin{footnotesize}
\begin{longtable}{@{}p{2.6cm} p{1.8cm} p{2.6cm} p{2.6cm} p{3.2cm}@{}}
\caption{Provenance, licensing, and terms of use of all third-party assets used in this work, and licenses of the artifacts we release. Citations are given at first use in the main text.}
\label{tab:licenses}\\
\toprule
\textbf{Asset} & \textbf{Version / accessed} & \textbf{Source} & \textbf{License / terms} & \textbf{Use \& compliance} \\
\midrule
\endfirsthead
\multicolumn{5}{l}{\emph{Table~\ref{tab:licenses} continued from previous page}}\\
\toprule
\textbf{Asset} & \textbf{Version / accessed} & \textbf{Source} & \textbf{License / terms} & \textbf{Use \& compliance} \\
\midrule
\endhead
\midrule
\multicolumn{5}{r}{\emph{Continued on next page}}\\
\endfoot
\bottomrule
\endlastfoot

\multicolumn{5}{@{}l}{\textbf{\emph{Datasets}}}\\
\hline
\addlinespace[2pt]
DILIrank / DILIrank~2.0 \citep{Olubamiwa2025}
  & 2.0
  & FDA LTKB portal\footnotemark[1]
  & U.S.~FDA / NCTR; U.S.~Government work, generally not subject to U.S.~copyright
  & Ground-truth DILI labels (Sec.~\ref{section:datasets}); attributed to FDA. \\
DILIst \citep{Thakkar2020}
  & 2020 release
  & Supplementary material of cited article
  & 
  & Ground-truth labels via the DILI Predictor split. \\
DILI Predictor split \& processed compounds \citep{Seal2024}
  &
  & GitHub release of the cited paper
  &
  & Train/test partition reused for benchmark comparability. \\
Garcia de Lomana liver \emph{in vitro} endpoints \citep{GarciadeLomana2025}
  & 2025
  & Supplementary data of cited article
  &
  & Proxy mechanistic endpoints in HADES (Sec.~\ref{section:datasets}). \\
ChEMBL \citep{Mendez2018}
  & 
  & \url{https://www.ebi.ac.uk/chembl/}
  & CC BY-SA 3.0 Unported
  & IC$_{50}$/EC$_{50}$/$K_d$ data for proxy MIE collection; ShareAlike obligation respected by releasing derived data under a compatible licence. \\
EveBIO \citep{evebio}
  & Release $9$
  & \url{https://evebio.org/}
  & Creative Commons CC BY-NC-SA 4.0 license
  & Complementary bioactivity records for proxy MIE collection. \\
AOP-Wiki \citep{AOPWiki}
  & accessed 2026-04-21
  & \url{http://aopwiki.org}
  & Creative Commons Attribution-ShareAlike (CC BY-SA)
  & Scaffold of MIEs, KEs, and AOPs underlying HADES. \\
PubChem
  & accessed 2026-04-21
  & \url{https://pubchem.ncbi.nlm.nih.gov/}
  & NCBI Website and Data Usage Policies; data not subject to U.S.~copyright
  & Synonym verification in TxGemma recognition audit (App.~I). \\
LiverTox \citep{livertox}
  & 2012, ongoing
  & \url{https://www.ncbi.nlm.nih.gov/books/NBK547852/}
  & NIDDK / U.S.~Government work
  & A--E likelihood scale reused descriptively, with attribution. \\
\addlinespace[4pt]
\hline
\multicolumn{5}{@{}l}{\textbf{\emph{Models}}}\\
\hline
\addlinespace[2pt]
Boltz-1, Boltz-2 \citep{Wohlwend2024, Passaro2025}
  & latest as of 2026-04-21
  & \url{https://github.com/jwohlwend/boltz}
  & MIT (code \& weights)
  & Trunk embeddings for AOP Predictor, Similarity Search, affinity gating. \\
TxGemma-27B-Predict / -Chat \citep{TxGemma}
  & 27B
  & HuggingFace \texttt{google/txgemma-27b-*}
  & Health AI Developer Foundations Terms of Use\footnotemark[2]
  & Non-clinical research benchmarking; HAI-DEF Clinical Use restriction not triggered. \\
Gemini-3-Flash-Preview, Gemini-3.1-Pro-Preview
  & accessed 2026-04-21
  & Gemini API (Google DeepMind)
  & Google Generative AI APIs Additional Terms of Service\footnotemark[3]
  & HADES backbone, LLM-as-judge, baseline (App.~J); use within research benchmarking clauses. \\
GPT-5.5
  & accessed 2026-04-21
  & OpenAI API
  & OpenAI Business Terms / Usage Policies
  & Non-commercial research benchmarking baseline (App.~J). \\
DILI Predictor \citep{Seal2024}
  & released version
  & \url{https://github.com/srijitseal/DILI_Predictor}
  & MIT License
  & External baseline. \\
BioTransformer 3.0 \citep{Wishart2022}
  & 3.0
  & \url{https://bitbucket.org/wishartlab/biotransformer3.0jar/}
  & GNU GPL v3 (envimicro module additionally relies on EnviPath CC BY-NC-SA 4.0 data; not used here)
  & Human metabolite enumeration (App.~\ref{app:tool_implementation:metabolites}); only the human modules are invoked. \\
BoLEK (\citep{Bolek})
  & \rule{1.5cm}{0.4pt}
  & \rule{2.4cm}{0.4pt}
  & \rule{1.8cm}{0.4pt}
  & Structural description module. \\
\addlinespace[4pt]
\multicolumn{5}{@{}l}{\textbf{\emph{Software libraries}}}\\
\addlinespace[2pt]
LangGraph \citep{LangGraph}
  & 1.1.10
  & \url{https://github.com/langchain-ai/langgraph}
  & MIT
  & State-machine orchestration of HADES. \\
DeepEval / G-Eval \citep{GEval}
  & 3.9.9
  & \url{https://github.com/confident-ai/deepeval}
  & Apache License 2.0
  & Judge harness for G-Eval metric (App.~G). \\
RDKit
  & 2026.3.1
  & \url{https://www.rdkit.org/}
  & BSD 3-Clause
  & Cheminformatics standardisation (App.~A). \\
\end{longtable}
\end{footnotesize}

\footnotetext[1]{\url{https://www.fda.gov/science-research/liver-toxicity-knowledge-base-ltkb/drug-induced-liver-injury-rank-dilirank-20-dataset}}
\footnotetext[2]{\url{https://developers.google.com/health-ai-developer-foundations/terms}}
\footnotetext[3]{\url{https://ai.google.dev/gemini-api/terms}}


\end{document}